\documentclass[journal]{IEEEtran}
\IEEEoverridecommandlockouts                              

\usepackage{url}
\usepackage{times}

\usepackage{graphicx}
\usepackage{subcaption}
\usepackage{booktabs}
\usepackage{hyperref}
\usepackage{makecell}
\usepackage{multirow}

\usepackage{pdfrender}
\usepackage{lscape}
\usepackage{amssymb}
\usepackage{amsmath}

\usepackage[capitalize]{cleveref}

\usepackage[firstpageonly=true]{draftwatermark}

\begin{document}

\SetWatermarkAngle{0}
\SetWatermarkColor{black}
\SetWatermarkLightness{0.5}
\SetWatermarkFontSize{9pt}
\SetWatermarkVerCenter{30pt}
\SetWatermarkText{\parbox{30cm}{%
\centering This is the authors' final version of the manuscript published as:\\
\centering J. Rozlivek, A. Roncone, U. Pattacini \& M. Hoffmann,\\
``HARMONIOUS – Human-like reactive motion control and multimodal perception for humanoid robots,'' \\
\centering in IEEE Transactions on Robotics, 2024, (C) IEEE, \url{https://doi.org/10.1109/TRO.2024.3502216}. \\
\centering An accompanying video is available at \url{https://youtu.be/gw8JB-1R3bs}.
}}

\title{\LARGE \bf HARMONIOUS -- Human-like reactive motion control and multimodal perception for humanoid robots}

\author{Jakub Rozlivek, Alessandro Roncone, Ugo Pattacini, Matej Hoffmann
\thanks{
J.R. and M.H. were supported by the Czech Science Foundation (GA CR), project no. 20-24186X. A.R. was supported by the National Science Foundation (NSF) grant No 2222952/2953. 
The authors thank Nicola Piga for the orientation sampling code.
The source code of the controller is available on GitHub (\url{https://github.com/robotology/react-control}).}
\thanks{Corresponding author: Matej Hoffmann ({\tt matej.hoffmann@fel.cvut.cz)}.}
\thanks{Jakub Rozlivek and Matej Hoffmann are with Department of Cybernetics, Faculty of Electrical Engineering, Czech Technical University in Prague, Czech Republic.} \thanks{Alessandro Roncone is with Human Interaction and RObotics (HIRO), Department of Computer Science, University of Colorado Boulder, Boulder, CO 80309 USA.} 
\thanks{Ugo Pattacini is with iCub Tech, Istituto Italiano di Tecnologia, 16163 Genova, Italy.}
}

\maketitle

\begin{abstract}
For safe and effective operation of humanoid robots in human-populated environments, the problem of commanding a large number of Degrees of Freedom (DoF) while simultaneously considering dynamic obstacles and human proximity has still not been solved. We present a new reactive motion controller that commands two arms of a humanoid robot and three torso joints (17 DoF in total). We formulate a quadratic program that seeks joint velocity commands respecting multiple constraints while minimizing the magnitude of the velocities. We introduce a new unified treatment of obstacles that dynamically maps visual and proximity (pre-collision) and tactile (post-collision) obstacles as additional constraints to the motion controller, in a distributed fashion over the surface of the upper body of the iCub robot (with 2000 pressure-sensitive receptors). 
This results in a bio-inspired controller that: 
(i) gives rise to a robot with whole-body visuo-tactile awareness, resembling peripersonal space representations, and (ii) produces human-like minimum jerk movement profiles. 
The controller was extensively experimentally validated, including a physical human-robot interaction scenario.
\end{abstract}

\IEEEpeerreviewmaketitle


\section{Introduction}
As robots are leaving safety fences and starting to share workspaces and even living spaces with humans,
they need to function in dynamic and unpredictable environments. Highly redundant platforms like humanoid robots have the possibility of performing tasks even in the presence of many constraints and obstacles. Given the dynamic nature of obstacles and real-time constraints, reactive motion control rather than planning is the solution. The key to success is whole-body awareness drawing on dynamic fusion of multimodal sensory information, inspired by peripersonal space representations in humans.

\begin{figure}[!t]
    \centering
    \begin{subfigure}[b]{0.48\textwidth}
    \centering
        \includegraphics[width=0.85\textwidth]{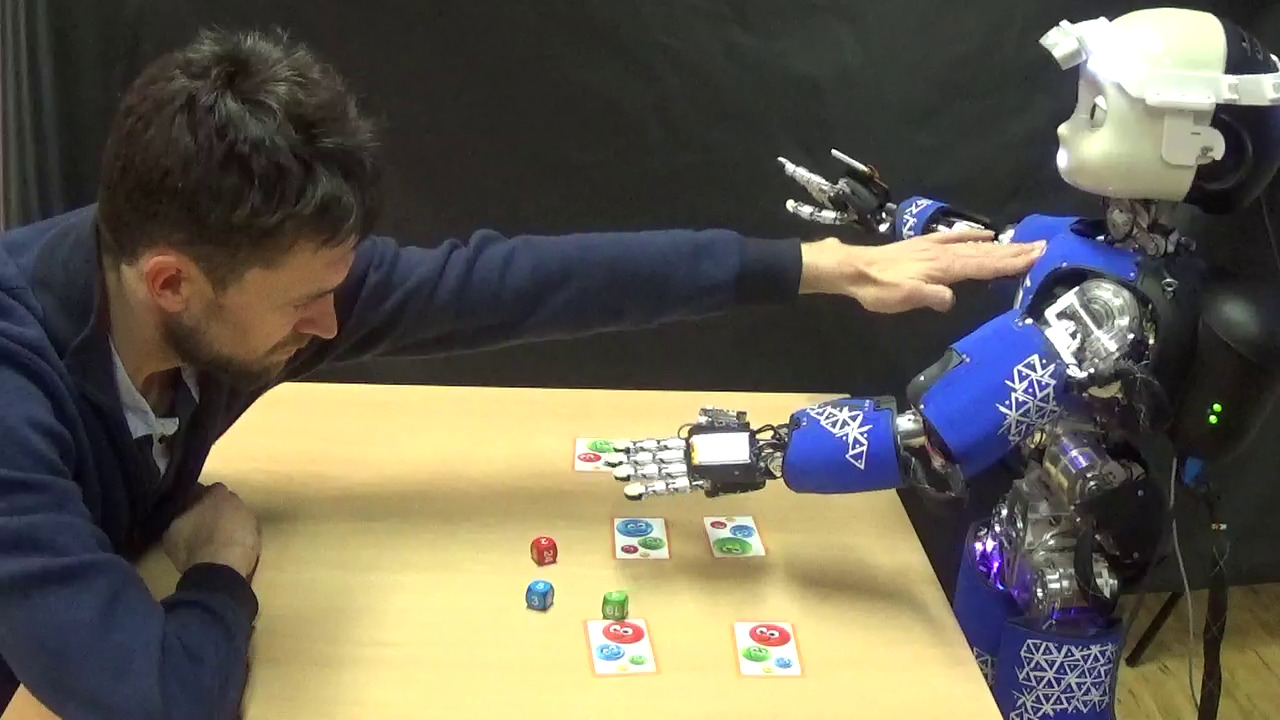}
        \caption{The robot is approaching the correct card while back bending to dodge the collision.}
    \end{subfigure}
    
    \begin{subfigure}[b]{0.48\textwidth}
    \centering
        \includegraphics[width=0.85\textwidth]{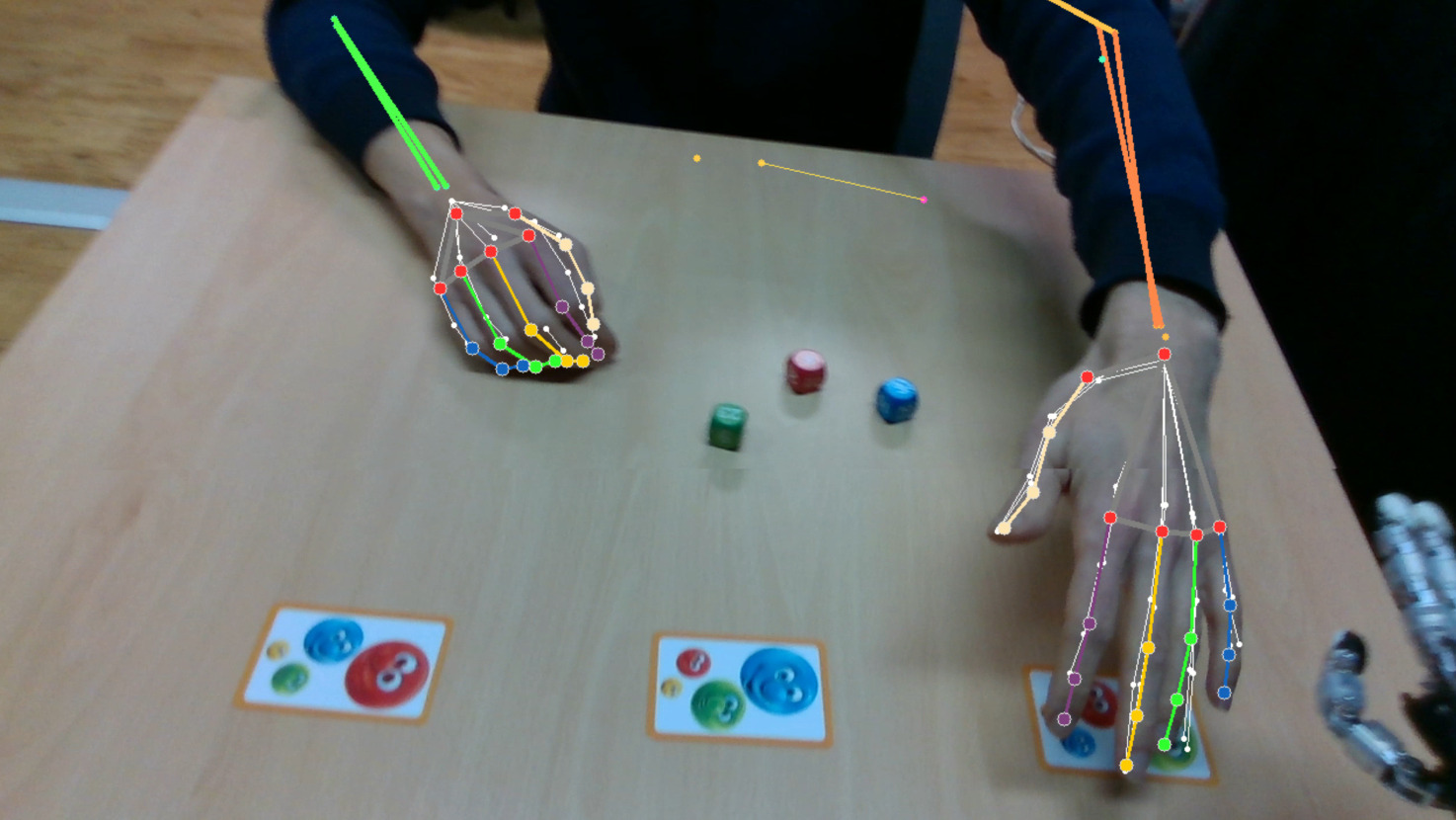}
        \caption{The robot is keeping a safe distance from the opponent's detected hand.}
    \end{subfigure}
    \caption{iCub robot in action during an interactive game demonstration.}
    \label{fig:intro_photo}
\end{figure}

We present a reactive motion controller that commands the upper body of the iCub humanoid robot ($2\times7$ Degrees of Freedom (DoFs) in every arm; $3$ torso DoFs), whereby the two hands can have separate tasks in Cartesian space or a common task (see Fig.\ref{fig:intro_photo}). Unique to our approach, obstacles perceived in three different modalities---visual and proximity (pre-collision) and tactile (post-collision)---are dynamically aggregated and projected first onto locations on individual body parts and then remapped as constraints into the joint velocity space. 

\textbf{Contributions.}
We have developed \textsl{HARMONIOUS}, a real-time reactive motion control system for upper body control of a humanoid robot sensing and avoiding contact with humans in its close proximity. Our specific contributions are the following.
\begin{enumerate}
\item \textsl{HARMONIOUS} is human-like in two important aspects: (i) it employs multimodal sensing around the whole body of the robot, ``visuo-tactile awareness'' resembling peripersonal space representations; (ii) it produces minimum jerk movement profiles which are characteristic of human motion. Together, these provide a basis for safe and natural human-robot interaction (HRI)---physical and social. 

\item We developed a unified representation of the space around the robot that feeds a reactive motion controller. Dynamically moving obstacles ranging from physical contact with the robot (zero distance, perceived through touch) to obstacles in close or far proximity can be remapped onto the robot's body parts, weighted, and transformed into kinematic constraints. Thus, there is a dense and multimodal representation of a peripersonal space analogue, which we refer to as the ``perirobot space'' (see \cite{rozlivek2023perirobot} for more details). It is demonstrated here by combining touch, proximity and visual sensors, but other contact-related or range-based sensors could be used.

\item We have extensively and experimentally evaluated \textsl{HARMONIOUS} and demonstrated its superior performance to state-of-the-art robot controllers where possible (\cite{haviland_neo_2021,nguyen_merging_2018,nguyen_compact_2018}). Moreover, to our knowledge, this is the first work to show real-time control of a humanoid robot upper torso (in total $17$ DoFs, with the possibility of different tasks for the two arms) faced with tens of dynamically moving obstacles around the whole robot body surface. Furthermore, kinematic singularities are handled through velocity damping and preferred postures are rewarded in the optimization problem formulation (see Tab.~\ref{tab:rel_work} for an overview). As the final evaluation, we tested \textsl{HARMONIOUS} in an interactive board game scenario with the human player dynamically perceived, simultaneously processing keypoints on his body, proximity signals, and physical contacts on the whole robot body.

\item The problem formulation is highly modular and both the minimization criteria (e.g., motivating preferred postures) or constraints to the quadratic program (e.g., obstacles) can be easily removed or added on the run.
\end{enumerate}

\section{Related work}
\subsection{Inverse kinematics}
Robot tasks are defined in \textsl{Cartesian} (also called \textsl{operational} or \textsl{task}) \textsl{space} while they are commanded in joint space, some form of \textsl{inverse kinematics (IK)}  mapping ($\mathbf{q} = f(\mathbf{x})$) is indispensable.
For industrial manipulators, a closed-form analytical solution is often available, especially for 6 DoF arms with a so-called spherical wrist. Closed-form solutions are global and fast to compute. However, such analytical solutions exploit specific geometric relationships and are generally not available for redundant robots with an arbitrary kinematic structure (\cite{park2012} is an exception). 
Numerical approaches to IK come with a higher computational load, but are more general and applicable to any kinematic chain. TRAC-IK~\cite{beeson_tracik_2015} is a state-of-the-art IK solver that solves IK as a sequential quadratic programming problem. Lloyd et al.~\cite{lloyd_fast_2022} presented an IK algorithm that uses the third-order root finding method to increase the speed and robustness of finding IK solutions. The solution consists of a set of target joint angles only---smooth trajectory generation and execution have to be taken care of separately. Pattacini et al.~\cite{pattacini_experimental_2010} combined a trajectory generator, a non-linear IK solver, and a Cartesian controller to ensure human-like minimum jerk movements in both Cartesian and joint space. None of the above methods can handle dynamic obstacles in the workspace. 
An overview of the related work from this and the following sections is in Tab.~\ref{tab:rel_work}. 

\begin{table*}[tb]
\centering
\resizebox{500pt}{!}{%
\begin{tabular}{l|lllrllll}
\textbf{Work} & \textbf{Control problem} & \textbf{\begin{tabular}[c]{@{}l@{}}React. obs. \\ avoidance\end{tabular}} & \textbf{\begin{tabular}[c]{@{}l@{}}Sensory modalities \\ to perceive obstacles\end{tabular}} & \textbf{\begin{tabular}[c]{@{}l@{}}DoF \\ shown\end{tabular}} & \textbf{\begin{tabular}[c]{@{}l@{}}Singularities \\ handling\end{tabular}} & \textbf{\begin{tabular}[c]{@{}l@{}}Pref. posture \\ motivation\end{tabular}}& \textbf{\begin{tabular}[c]{@{}l@{}}Ori. error \\ minim.\end{tabular}} & \textbf{\begin{tabular}[c]{@{}l@{}}Explicit \\ traj samp.\end{tabular}} \\ \hline
\textbf{Ours} & DiffK - QP problem & Lin. ineq. constr & 3 - vision, proximity, touch & 17 & Velocity damping & Yes & Yes & Yes \\
\cite{beeson_tracik_2015} & IK - Sequential QP & - & - & 15 & No & No & Yes & No \\
\cite{lloyd_fast_2022} & IK - Jacobian based & - & - & 16 & Damped variant & No & Yes & No \\
\cite{pattacini_experimental_2010} & IK - Nonlinear opt. & - & - & 7 & No & Yes & Yes & Yes \\
\cite{glass_realtime_1995} & DiffK - Damped LS & Lin. ineq. constr. & 0 - known obstacles & 8 & Damped LS & No & Yes & No \\
\cite{park_movement_2008} & DiffK - Pseudoinverse & Potential field & 0 - known obstacles & 7 & No & No & Yes & Yes \\
\cite{flacco_depth_2012} & DiffK - Pseudoinverse & Repulsive vector & 1 - vision & 7 & No & No & No & No \\
\cite{haviland_neo_2021} & DiffK - QP problem & Lin. ineq. constr. & 0 - known obstacles & 7 & Max. manipulability & No & Yes & No \\
\cite{escobedo_contact_2021} & DiffK - QP problem & Lin. ineq. constr. & 2 - proximity, touch & 7 & Velocity damping & Yes & No & No \\
\cite{ding_collision_2020} & DiffK - QP problem & Lin. ineq. constr. & 1 - proximity & 7 & Velocity damping & No & No & No \\
\cite{guo_new_2012} & DiffK - QP problem & Lin. ineq. constr. & 0 - known obstacles & 7 & No & No & Yes & No \\
\cite{tong2023} & DiffK - QP problem & - & - & 12 & Max. manipulability & No & Yes & No \\
\cite{suleiman_inverse_2016} & DiffK - QP problem & Lin. ineq. constr. &  0 - known obstacles & 6 & No & No & Yes & No \\
\cite{rakita2021} & IK - Nonlinear opt. & Cost function & 0 - known obstacles & 8 & Max. condition number & No & Yes & No\\
\cite{khatib_realtime_1986} & Operational space control & Potential field & 0 - known obstacles & 6 & No & No & Yes & No \\
\cite{merckaert_realtime_2022} & ERG + PD control & Potential field & 1 - vision & 7 & No & No & Yes & No \\
\cite{nguyen_merging_2018} & IK - Nonlinear opt. & Repulsive vector & 2 - vision, touch & 10 & No & No & Yes & Yes\\
\end{tabular}
}
\caption{Comparison of different approaches to solving the inverse kinematics problem. Control problem acronyms are: differential kinematics (DiffK), inverse kinematics (IK), quadratic programming (QP), least squares (LS), and explicit reference governor (ERG).
\label{tab:rel_work}}
\end{table*}

\subsection{Obstacle avoidance and differential kinematics}
Collision avoidance in static environments can be resolved at a higher level by using trajectory optimization for robot motion planning \cite{schulman_finding_2013,zimmermann_differentiable_2022,bordalba2023}. However, currently, these methods cannot provide solutions for scenarios like humanoid robots interacting with dynamic human-populated environments in real time. 
The family of methods that alleviate the above-mentioned problems and that are gaining popularity recently rely on differential kinematics, i.e. the formulation of the operational space task in velocity space and a mapping between desired Cartesian velocity and joint velocity through the robot Jacobian. Differential kinematics, also referred to as Resolved-rate motion control (RRMC), dates back to Whitney~\cite{whitney_resolved_1969}. This method is inherently local---depends on the current robot configuration and the corresponding Jacobian---and as it provides joint velocities, it can be directly employed for motion control (without separate trajectory generation). 

Inverse differential kinematics methods rely on some form of Jacobian inverse ($\mathbf{\dot{q}}= \mathbf{J}^{-1}(\mathbf{q})\mathbf{\dot{x}}$), more often pseudo-inverse when $J$ is not square, like for redundant robots.  The damped least squares (DLS) method \cite{glass_realtime_1995,park_movement_2008,flacco_depth_2012} is preferred over Jacobian pseudo-inverse or transpose as it ``dampens'' the motions in the vicinity of kinematic singularities. To dodge obstacles in redundant robots, self-motion, i.e. the possibility to use the \textsl{null space} of the solution for additional tasks, is exploited \cite{albini_exploiting_2021,cirillo_conformable_2016,magrini_hybrid_2016, mansard2009}. The advantage of these approaches is that the problem has a clear hierarchical structure that allows to set priorities. For example, Albini et al.~\cite{albini_exploiting_2021} define obstacle avoidance as the primary task and the reaching target uses the null space.
The disadvantage is the limited flexibility of the problem formulation in face of many (some dynamically appearing) constraints or their changing priorities.
 
Alternatively, the motion control task can be defined using the forward differential formulation, searching for joint velocities $\mathbf{\dot{q}}$ that best match the desired Cartesian velocity of the end effector $\mathbf{\dot{x}}$, thus minimizing $||\mathbf{\dot{x}} - \mathbf{J}(\mathbf{q})\mathbf{\dot{q}}||$. To this end, it is possible to use quadratic programming (QP) optimization with a quadratic cost function and linear constraints \cite{haviland_neo_2021,escobedo_contact_2021,ding_collision_2020,guo_new_2012,tong2023,suleiman_inverse_2016,escobedo_framework_2022}.
To avoid singular configurations, two options are usually given: add manipulability maximization to the cost function \cite{haviland_neo_2021,tong2023} or dampen joint velocities when close to singularities \cite{escobedo_contact_2021,ding_collision_2020,nakamura1986}.
In addition, undesirable configurations can be avoided by extending the optimization cost function to motivate the robot to be close to the middle of the joint limits \cite{escobedo_contact_2021} or close to the robot resting position \cite{pattacini_experimental_2010}.
In the case of the QP problem, obstacles are represented as linear inequality constraints allowing motions in tangential directions improving target reachability \cite{haviland_neo_2021,escobedo_contact_2021, ding_collision_2020}.

Most of the mentioned works do not handle self-collisions explicitly. 
The exceptions are \cite{ding_collision_2020}, where self-collisions are solved by adding additional obstacle points for robot links and the end effector, and \cite{rakita2021}, where a neural network is trained to favor configurations far from self-collision states. 
Moreover, De Santis et al. \cite{desantis2007} presented an algorithm for real-time generation of self-collsion avoidance motions using repulsion forces converted to avoidance torques.
Unlike most of the mentioned controllers that presented solutions of the IK problem for a single-arm robot, Tong et al. \cite{tong2023} proposed a four-criterion-optimization coordination motion scheme for a dual-arm robot that deals with coordination constraints and physical constraints. 
However, their approach does not consider obstacle avoidance and the torso joints of the dual-arm robot. 

To represent obstacles, artificial potential fields \cite{khatib_realtime_1986}, where obstacles are repulsive surfaces that repel the robot from the obstacle, can be used \cite{merckaert_realtime_2022,dietrich_reactive_2012,haddadin_realtime_2010}. 
Park et al. \cite{park_movement_2008} presented a dynamic potential field for smoother avoidance movements.
Repulsive vectors in Cartesian space are shown in \cite{flacco_depth_2012} and later used in \cite{nguyen_merging_2018,nguyen_compact_2018}.

\subsection{Human-like motion generation}
Several controllers can work as explicit local trajectory generators to sample the trajectory between the start and the target pose. 
The algorithm in \cite{haddadin_realtime_2010} combines control with path planning and can handle arbitrary desired velocity profiles for the robot. 
The Cartesian controller \cite{pattacini_experimental_2010} produces human-like quasi-straight trajectories with minimum-jerk profiles characteristic of human motions.
A reactive controller in \cite{nguyen_merging_2018,nguyen_compact_2018} uses the minimum-jerk trajectory sampling as in \cite{pattacini_experimental_2010}. 
Suleiman \cite{suleiman_inverse_2016} proposed an algorithm for implicit minimum jerk trajectories by replacing joint velocities by joint jerk as control parameters in the optimization problem.

\subsection{Sensing of obstacles and contacts}
To make safe movement of a robot in a cluttered or human-populated environment possible, obstacles need to be perceived in real time and processed either before a collision happens or immediately after contact with the robot. We will focus on perception in the context of HRI.

To sense at a distance, RGB-D cameras are frequently adopted \cite{flacco_depth_2012,lambert_joint_2019,he_visibility_2022,kulic_realtime_2006,docekal2022}. Depth information can be used to compute a set of spheres that represent obstacles \cite{kulic_realtime_2006} or to create a depth space representation that is used to generate repulsive vectors for the robot \cite{flacco_depth_2012}. 
Magrini and De Luca~\cite{magrini_hybrid_2016} estimated the contact force based on the contact point detected by a Kinect sensor. 
Alternative visual sensing for obstacle detection are tracking systems that follow markers attached to specific body parts, e.g., wrist \cite{haddadin_realtime_2010} or arm joints and head \cite{merckaert_realtime_2022}.
Aljalbout et al. \cite{aljalbout_learning_2020} used a static RGB camera in simulation and trained a convolutional neural network for obstacle avoidance.

Collisions with a moving robot may be acceptable provided that the contact forces are within limits. Collisions need to be detected, located, identified (e.g., determining the forces), and, possibly following classification of the collision, reacted upon (e.g., stop, slow down, retract, etc.) \cite{Haddadin2017}. Force/torque sensors within the robot structure, together with models of its dynamics, can be employed. Alternatively, contacts can be perceived through tactile sensors. We will focus on large-area sensitive skins. Unlike when force/torque sensors are employed, collision localization/isolation is easier, even for multi-contacts. Cirillo et al. \cite{cirillo_conformable_2016} developed a flexible skin, allowing measurement of contact position and three components of contact force. Calandra et al. \cite{calandra_learning_2015} presented a data-driven method for a fast and accurate prediction of joint torques from contacts detected by tactile skin. 
Albini et al. used robot skin feedback for HRI \cite{albini_enabling_2017} as well as robot control in a cluttered environment \cite{albini_exploiting_2021}. 
Kuehn and Haddadin~\cite{kuehn_artificial_2017} introduced an artificial robot nervous system, which integrates tactile sensation and reflex reactions into robot control through the concept of robot pain sensation. Some large-area sensitive skins have been safety-rated and are deployed in collaborative robot applications (see Airskin and the analysis of its effects in \cite{Svarny2022Airskin}). 

Individual visual sensors---on the robot itself or in other locations in the workspace---are prone to occlusions (which can be mitigated by employing multiple sensors \cite{flacco2016real}). Large area tactile arrays are distributed over the whole robot body, but only respond after physical contact. Distributed proximity sensors provide an alternative: sensing at a small distance over the whole robot body. Proximity \cite{escobedo_contact_2021,ding_collision_2020} or tomographic \cite{muhlbacher-karrer_contactless_2017} sensors can be used for anticipation of collisions and contactless motion guidance.

Typically, only one sensor type is used, and the corresponding robot control is more or less dependent on the nature of the sensory information. 
On the other hand, there is a field of multimodal sensor fusion. Dean-Leon et al.~\cite{deanleon2016} presented a fusion of pressure and proximity data from multimodal artificial skin to change the robot's dynamic behavior, e.g., making a stiff robotic system compliant. Hierarchical control is often used together with the integration of several sensors, such as an artificial skin for collision avoidance with a camera for position-based visual servoing \cite{martin2020}, or a LIDAR for collision avoidance and a multimodal skin for tactile interaction or distance/force control \cite{armleder2022}. In addition to collision avoidance, fusion can be used for other tasks, such as opening the door \cite{wieland2009}. Multimodal fusion can integrate more than sensor data as in \cite{felip2014}, where contact location is estimated using a contact hypotheses fusion approach from multiple sensor modalities, the context of interaction, and the environment. 

Humans posses a dynamically established protective safety margin around the whole body, drawing on visuo-tactile interactions: Peripersonal space (PPS), e.g., \cite{clery2018frontier}. Roncone et al.~\cite{roncone2016, roncone2015} took inspiration from PPS and implemented visuo-tactile receptive fields on a humanoid robot. Nguyen et al.~\cite{nguyen_merging_2018,nguyen_compact_2018} deployed this in HRI to feed a reactive motion controller. As affordable and increasingly accurate sensors of different types are becoming ubiquitous, it is important and timely to develop a unified treatment of their signals such that they can be processed by a single motion controller (see also \cite{rozlivek2023perirobot} where a unified treatment of the perirobot space for collaborative robot applications is proposed). 

\subsection{Comparison of \textsl{HARMONIOUS} with related works}
In this work, we present a framework in which RGB-D, proximity, and tactile sensors are simultaneously processed, related to the complete robot body, and used to generate real-time constraints for a whole-body reactive motion controller. 

To the extent that this is possible, the main components of our solution are categorized and contrasted with representative works from the literature in Tab.~\ref{tab:rel_work}. 

We formulate the control problem (second column in Tab.~\ref{tab:rel_work}) as a quadratic program in velocity space---employing differential kinematics, similarly to many recent works in this area \cite{haviland_neo_2021,escobedo_contact_2021,ding_collision_2020,guo_new_2012,tong2023,suleiman_inverse_2016}. Reactive obstacle avoidance (second column) is implemented using a linear inequality constraint (LIC). Importantly, this work is to our knowledge the first to employ three different sensory modalities (vision, proximity, touch), in a unified manner, to perceive and avoid obstacles (fourth column). The fifth column (DoF shown) lists the number of DoF employed in our approach and in the literature. \textsl{HARMONIOUS} handles kinematic singularities through velocity damping. The robot is motivated to favor joint positions near a preferred posture, e.g., the center of the joint ranges or a home configuration to prevent undesirable configurations, unlike most of the works in the literature. \textsl{HARMONIOUS} can handle position and orientation targets (Orientation error minimization) and samples the trajectory of end-effector positions (Explicit trajectory sampling), inspired by human reaching movements---quasi-straight trajectories with bell-shaped velocity profiles. 
The end-effector orientation sampling is calculated using a computer graphics method \textsl{slerp} for constant angular velocities, ensuring smooth rotations and the shortest path between initial and final orientations.
In summary, \textsl{HARMONIOUS} employs state-of-the-art control problem and reactive obstacle avoidance formulation, but moves beyond the state of the art in providing a unified treatment of three different sensory modalities, the number of DoFs controlled, and incorporation of preferred position motivation and orientation error minimization.
Beyond what could fit in Tab.~\ref{tab:rel_work}, \textsl{HARMONIOUS} also features self-collision avoidance and the possibility of setting targets (tasks) for the two robot arms independently or jointly (bimanual task). To our knowledge, \textsl{HARMONIOUS} is unparalleled in the scale of the problem and the number of features it can tackle.

\section{Materials and Methods}
\begin{figure*}[htbp]
    \centering
    \includegraphics[width=\textwidth]{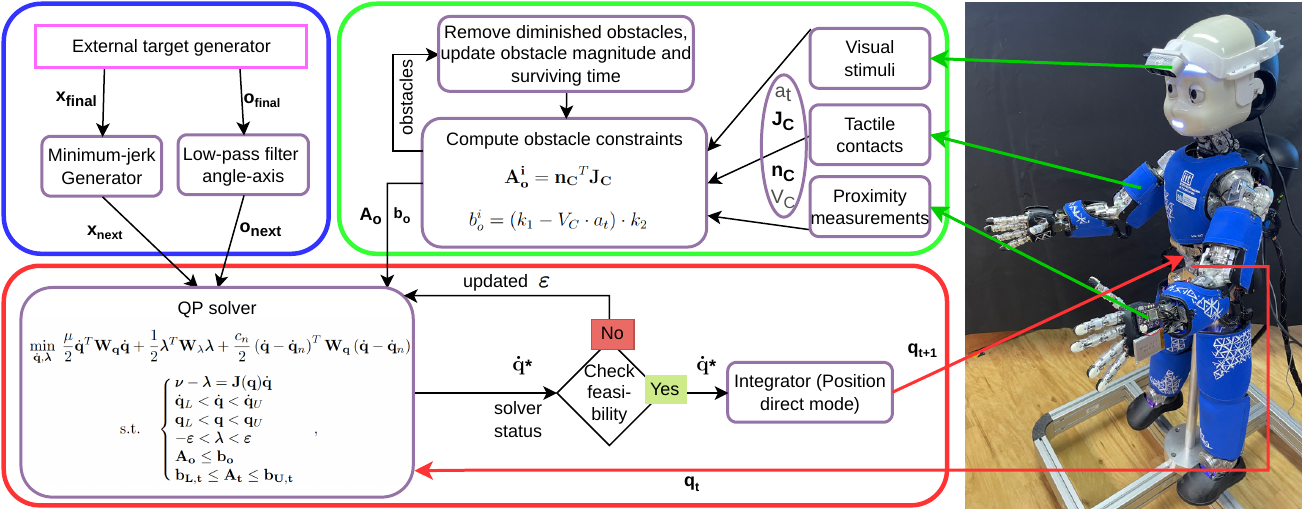}
\caption{\textsl{HARMONIOUS} -- overview. The blue box represents the local trajectory sampling. The trajectory towards the target pose is sampled using a minimum-jerk generator (position) and a low-pass filter (orientation). The target pose is received from an external target generator. The green box represents the obstacle processing. It takes inputs from proximity sensors, skin parts (tactile stimuli), and RGB-D camera (visual stimuli) and computes linear inequality constraints to dodge obstacles or react to the collision that occurred. The obstacle exists virtually for a specific surviving time with a decreasing threat level, giving humans time to react. The red box represents the controller with a QP solver that solves the differential kinematics problem. It takes the next desired pose and current joint positions as input. If the problem is feasible, the computed joint velocities are integrated into new joint positions and sent to the robot. Otherwise, the position constraint is relaxed, and the QP problem is solved again.}
    \label{fig:overview}
\end{figure*}

Here we present \textsl{HARMONIOUS} -- Human-like reactive motion control and multimodal perception for humanoid robots. An overview is provided in Fig.~\ref{fig:overview}. The core of the solution is the motion controller (red box) and the actual QP formulation (magenta) which is online fed by the target trajectory generator (blue box) and the constraints projected from obstacles (green box).
\subsection{Experimental setup}
\label{subsec:setup}
The experiments were conducted with the humanoid robot iCub (\cite{icub}, see \cref{fig:intro_photo}), specifically on its upper body with 17 degrees of freedom (DoF) consisting of two 7-DoF arms and a common 3-DoF torso. 
The robot was equipped with an RGB-D camera Intel RealSense D435 placed above his eyes to track obstacles in 3D, an artificial electronic skin \cite{icubSkin} to detect contacts and their intensity, and two proximity sensor units \cite{watanabe2021} with time-of-flight sensors that improve perception in ``blind spots'' (back of hands).

\subsection{Motion controller}
\label{subsec:controller}
The main component of our approach is a motion controller (see the red box in Fig.~\ref{fig:overview}).
Our motion controller is based on reactive kinematic control, in particular the approach called differential kinematics which is defined as
\begin{equation}
    \boldsymbol{\nu} = \mathbf{J}(\mathbf{q})\dot{\mathbf{q}}
\end{equation}
where $\boldsymbol{\nu} \in \mathbb{R}^{m}$ is the spatial velocity of the end effector ($m$ is the task dimension---6 for single-arm version, 12 for dual-arm version), $\dot{\mathbf{q}} \in \mathbb{R}^{n}$ are the joint velocities of the robot ($n$ is the number of joints) and  $\mathbf{J}(\mathbf{q}) \in \mathbb{R}^{m\times n}$ is the Jacobian of the robot for the joint positions $\mathbf{q} \in \mathbb{R}^{n}$. This relation represents a trajectory constraint between the robot configuration and Cartesian space. In this work, we assume the reference frame for Cartesian coordinates and Jacobians is the base frame of the robot (around the waist).

The spatial velocity of the end effector is computed from the current end-effector pose and the desired end-effector pose in the next time period as

\begin{equation}
\boldsymbol{\nu^{pos}} = \frac{\mathbf{x_t} - \mathbf{x_c}}{t_s}, \qquad \boldsymbol{\nu^{ori}} = \frac{\varphi(\mathbf{R_t}\mathbf{R_c}^T)}{t_s}
\end{equation}
where $\boldsymbol{\nu^{pos}} \in \mathbb{R}^{3}$ is the translational velocity of the end effector, $\mathbf{x_c} \in \mathbb{R}^{3}$ and $\mathbf{x_t} \in \mathbb{R}^{3}$ are the current and desired next end effector position, $\boldsymbol{\nu^{ori}} \in \mathbb{R}^{3}$ is the rotational velocity of the end effector, 
$\mathbf{R_c}\in \mathbb{R}^{3\times 3}$ and $\mathbf{R_t} \in \mathbb{R}^{3\times 3}$ are the current and desired next end effector orientation (converted from axis-angle representation), $\varphi$ represents mapping from a rotation matrix to an axis-angle representation to obtain angular velocities using a period time $t_s$ exploiting the infinitesimal rotation assumption (see e.g., \cite{haviland_neo_2021}).

For our purposes, we can relax the trajectory constraint (the ``reaching task'') by adding a slack variable $\boldsymbol{\lambda} \in \mathbb{R}^{m}$ to help the controller with finding a solution when other (e.g., safety) constraints appear:
\begin{equation} \label{eq:traj_constr}
    \boldsymbol{\nu} - \boldsymbol{\lambda} = \mathbf{J}(\mathbf{q})\dot{\mathbf{q}}
\end{equation}
Then, we formulate the problem as a strictly convex QP problem with joint velocities $\dot{\mathbf{q}}$ and slack variables $\boldsymbol{\lambda}$ as optimization variables. In our cost function, we minimize the sum of the squared Euclidean norm of the joint velocities, the distance from the preferred joint positions, and the squared Euclidean norm of the vector of slack variables. It is formulated as:
\begin{equation}
\label{eq:prob_formulation}
\min_{\dot{\mathbf{q}},\boldsymbol{\lambda}}\ \frac{\mu}{2}\dot{\mathbf{q}}^T\mathbf{W_q}\dot{\mathbf{q}} + \frac{1}{2}\boldsymbol{\lambda}^T\mathbf{W_\lambda}\boldsymbol{\lambda} + \frac{c_h}{2}\left(\dot{\mathbf{q}} - \dot{\mathbf{q}}_n \right)^T\mathbf{W_q}\left(\dot{\mathbf{q}} - \dot{\mathbf{q}}_n \right)
\end{equation}
\noindent
where diagonal matrices $\mathbf{W_q} \in \mathbb{R}^{n\times n}$ and $\mathbf{W_\lambda} \in \mathbb{R}^{m\times m}$ contain the weights of the individual joints and the relaxations for the trajectory constraint, $\mathbf{\dot{q}}_n$ are the velocities needed to reach the preferred (e.g., home) robot configuration, $c_h$ is a weight of the motivation to keep robot close to a preferred configuration, and $\mu$ is a damping factor calculated as:
\begin{equation}
    \mu = \begin{cases}
(1-\frac{\omega}{\omega_0})^2 + 0.01& \omega < \omega_0,\quad \omega_0 \text{ is a threshold,}  \\
0.01 & \text{otherwise.} 
\end{cases}
\end{equation}
where $\omega$ is a manipulability index \cite{yoshikawa1985} computed as $\omega=\sqrt{\mathrm{det}(\mathbf{J}(\mathbf{q})\mathbf{J(\mathbf{q})}^T)}=s_1 s_2 ...s_n$, where $s_i$ is the i-th singular value of $\mathbf{J}(\mathbf{q})$.
This damping factor increases when the manipulability index decreases to zero, implying that the robot is close to a singularity position (the manipulability index is zero in that position). The higher damping factor makes the minimization of joint velocities more important, leading to lower joint velocities and singularity avoidance~\cite{nakamura1986}. We assume that the weights ($\mathbf{W_q}$ $\mathbf{W_\lambda}$, $c_h$, and $\mu$) are non-negative to ensure that the problem is convex. The weights in $\mathbf{W_q}$ corresponding to the torso joints are higher than for the arm joints, which, through the minimization term, motivates smaller torso movements, which is more natural.

Apart from the mentioned relaxed trajectory constraint, the minimization is done with respect to several other constraints, such as joint velocity and position bounds (see Sec.~\ref{subsec:joint_constr}), slack variable bounds, collision avoidance (see Sec.~\ref{subsec:obstacles}), and other task or kinematic constraints (see Sec.~\ref{subsec:relpos_constr}). The constraints can be written as follows:
\begin{equation}
  \mathrm{s.t.} \quad  \begin{cases}
 \boldsymbol{\nu} - \boldsymbol{\lambda} = \mathbf{J}(\mathbf{q})\dot{\mathbf{q}} \\
  \dot{\mathbf{q}}_L < \dot{\mathbf{q}} < \dot{\mathbf{q}}_U \\
  \mathbf{q}_L < \mathbf{q} < \mathbf{q}_U \\
  -\boldsymbol{\varepsilon}  < \boldsymbol{\lambda}  < \boldsymbol{\varepsilon} \\
  \mathbf{A_o} \mathbf{\dot{q}} \leq \mathbf{b_o} \\
  \mathbf{b_{L,t}} \leq \mathbf{A_t} \mathbf{\dot{q}} \leq \mathbf{b_{U,t}}  \end{cases}
\end{equation}
where $\mathbf{\dot{q}}_{L,U} \in \mathbb{R}^{n}, \mathbf{q}_{L,U} \in \mathbb{R}^{n}, \boldsymbol{\varepsilon} \in \mathbb{R}^{m}$ are the lower and upper bounds for the joint velocities, the joint positions, and the slack variables, respectively. $\mathbf{A_o}\in \mathbb{R}^{m_o\times n}$ and $\mathbf{b_o} \in \mathbb{R}^{m_o}$ define ($m_o$) obstacle avoidance constraints; $\mathbf{A_t} \in \mathbb{R}^{m_t\times n}$, $\mathbf{b}_{L,t} \in \mathbb{R}^{m_t}$, and $\mathbf{b}_{U,t}\in \mathbb{R}^{m_t}$ define other ($m_t$) task constraints.

\subsection{Dual-arm trajectory constraint}
\label{subsec:dualarm_constraint}
Here, we formulate the problem for two arms with a common torso. Since we have a common torso of the robot, we cannot calculate the trajectory constraints for the arms independently of each other.
Moreover, it is not always possible to reach targets with both arms simultaneously. Conflict situations are managed by prioritizing one of the arms and calling it the ``primary arm'', and the other is then the ``secondary arm''. 
Therefore, the trajectory constraint (Eq.~\ref{eq:traj_constr}) for a robot with a common torso is expanded as:
\begin{equation}
\begin{split}
 \boldsymbol{\nu_p^{pos}} - \boldsymbol{\lambda^{pos}_p}& = \mathbf{J_{t,pos}^p}(\mathbf{q^t}) \mathbf{\dot{q}^t} +  \mathbf{J_{a,pos}^p}(\mathbf{q^p}) \mathbf{\dot{q}^p} \\
 \boldsymbol{\nu_p^{ori}} - \boldsymbol{\lambda^{ori}_p}& = \mathbf{J_{t,ori}^p}(\mathbf{q^t}) \mathbf{\dot{q}^t} +  \mathbf{J_{a,ori}^p}(\mathbf{q^p}) \mathbf{\dot{q}^p}  \\
 \boldsymbol{\nu_s^{pos}} - \boldsymbol{\lambda^{pos}_s}& = \mathbf{J_{t,pos}^s}(\mathbf{q^t}) \mathbf{\dot{q}^t} +  \mathbf{J_{a,pos}^s}(\mathbf{q^s}) \mathbf{\dot{q}^s}  \\
 \boldsymbol{\nu_s^{ori}} - \boldsymbol{\lambda^{ori}_s}& = \mathbf{J_{t,ori}^s}(\mathbf{q^t}) \mathbf{\dot{q}^t} +  \mathbf{J_{a,ori}^s}(\mathbf{q^s}) \mathbf{\dot{q}^s}
 \end{split}
\end{equation}
where the subscripts $t,p,s$ represent \textbf{t}orso, ``\textbf{p}rimary arm'', and ``\textbf{s}econdary arm'' respectively. The superscripts $pos, ori$ represent the relaxation of the constraint of the \textbf{pos}ition and \textbf{ori}entation trajectory, respectively. The single-arm formulation contains only the torso and the ``primary arm'' part. 
$\mathbf{J^{p/s}_{a,pos}}(\mathbf{q^{p/s}}) \in \mathbb{R}^{3\times (n-3)}, \mathbf{J^{p/s}_{t,pos}}(\mathbf{q^t}) \in \mathbb{R}^{3\times 3}$ are the translational velocity components of the Jacobian matrices for arm and torso, respectively, $\mathbf{J^{p/s}_{a,ori}}(\mathbf{q^{p/s}}) \in \mathbb{R}^{3\times (n-3)}, \mathbf{J^{p/s}_{t,ori}}(\mathbf{q^t}) \in \mathbb{R}^{3\times 3}$ are the rotational velocity components of the Jacobian matrices for arm and torso, respectively.

Consistent with the arm prioritization, we distinguish two cases for the limits of the slack variable $\boldsymbol\varepsilon$. They are set to $\infty$ (the deviation from the trajectory is minimized only) or to zero (the equality constraint without relaxation).
In our solution, $\boldsymbol\varepsilon$ is set to $\infty$ for all $\boldsymbol\lambda$ except $\boldsymbol{\lambda_p^{pos}} = [0,0,0]^T$.
If the equality constraint on the desired translation velocity of the ``primary arm'' makes the problem infeasible, we solve the problem again with the constraint relaxed to the minimization of the deviation. This strategy allows the robot to reach a target position with higher precision if the original problem is feasible and to make it feasible if the equality constraint is too strict. 

\subsection{Local trajectory sampling}
\label{subsec:local_sampling}
The first input to the reactive controller is the trajectory waypoint for the next time step (see the blue box in Fig.~\ref{fig:overview}). 
Given the desired final target pose from an external source (e.g., a high-level target generator), we compute the desired next-time pose separately in position and orientation. 
We use a low-pass filter in the axis-angle representation for orientation sampling, which ensures a constant-speed transition between two orientations. 
For position sampling, a generator of approximately minimum-jerk trajectories is employed. 
Position samples are generated using a third-order linear time-invariant dynamical system \cite{pattacini_experimental_2010}. 

We distinguish between streamed targets (e.g., reference tracking) and individual (reaching) targets. 
In the first case, the targets can be skipped (e.g., during collision avoidance), and only the newest one is important, and the local trajectory sampling is not applied as it is assumed that the movement smoothness is handled in the external target generator. 

\subsubsection{Position sampling}
\label{subsubsec:pos_sam}
The sampling of a time-varying input $\mathbf{x}_d \in \mathbb{R}^3$ expressing the desired positions for the end effector in Cartesian space is performed by means of a third-order filter, which is capable of generating output trajectories that exhibit the property of being quasi minimum-jerk, thus resembling the human-like movements as described in \cite{Flash1985}.

We tightly follow the implementation presented in \cite{pattacini_experimental_2010}, where the idea is to start from the feedback formulation of the minimum-jerk trajectory as the solution of the optimal problem proposed in \cite{shadmehr_wise_2005}. The exact solution resorts to a third-order linear time-variant (LTV) dynamical filter applied to each coordinate $x_d\left(*\right)$ of the vector $\mathbf{x}_d$:

\begin{equation} \label{eq:minjerk-ltv}
    \begin{bmatrix} \dot{x} \cr \ddot{x} \cr \dddot{x} \end{bmatrix} =
    \begin{bmatrix} 0 & 1 & 0 \\
                    0 & 0 & 1 \\
                    \frac{-60}{\left(T-t\right)^3} &
                    \frac{-36}{\left(T-t\right)^2} &
                    \frac{-9}{T-t} \end{bmatrix}
    \begin{bmatrix} x \cr \dot{x} \cr \ddot{x} \end{bmatrix} +
    \begin{bmatrix} 0 \cr 0 \cr \frac{60}{\left(T-t\right)^3} \end{bmatrix}
    x_d\left(*\right)
\end{equation}
where $T$ represents the execution time.

To circumvent the difficulty posed by coefficients becoming unbounded for $t \rightarrow T$, the authors ran in \cite{pattacini_experimental_2010} a second optimization to find out the third-order linear time-invariant (LTI) system that can deliver the best approximation of the output of Eq. \ref{eq:minjerk-ltv}, minimizing the jerk measure over the same temporal interval $\left[0,T\right]$.

The resulting LTI system is:
\begin{equation} \label{eq:minjerk-lti}
    \begin{bmatrix} \dot{x} \cr \ddot{x} \cr \dddot{x} \end{bmatrix} =
    \begin{bmatrix} 0 & 1 & 0 \cr
                    0 & 0 & 1 \cr
                    \frac{a}{T^3} & \frac{b}{T^2} & \frac{c}{T} \end{bmatrix}
    \begin{bmatrix} x \\ \dot{x} \\ \ddot{x} \end{bmatrix} +
    \begin{bmatrix} 0 \\ 0 \\ -\frac{a}{T^3} \end{bmatrix}
    x_d\left(*\right)
\end{equation}
where $a\approx-150.766$, $b\approx-84.981$, and $c\approx-15.967$ are the constant coefficients found by the optimization, and $T$ is a parameter (related to the execution time) that regulates the reactivity of the filter. Throughout our experiments, we set $T = \left(||\mathbf{x_d}-\mathbf{x_0}||/v_t\right)$ s, where $\mathbf{x_0} \in \mathbb{R}^3$ is the starting position, and $v_t$ is the desired Cartesian speed of movement. 

The output $\mathbf{x}\left(t\right) \in \mathbb{R}^3$ of the filter in Eq. \ref{eq:minjerk-lti} provides a sampling of the input $\mathbf{x}_d\left(t\right)$ that is ``quasi'' minimum-jerk, being the filter only an approximation of the system (Eq. \ref{eq:minjerk-ltv}), although the authors in \cite{pattacini_experimental_2010} proved that it is reliable and effective in smoothing out sharp transitions while guaranteeing bell-shaped velocity profiles.

\subsubsection{Orientation sampling}
\label{subsubsec:ori_sam}
We use the axis-angle representation for the orientations of the end effector. The orientation is described by a rotation vector $\mathbf{r} = \left[r_{1}, r_{2}, r_{3}\right]^T,$
that encodes the axis of rotation $u = \mathbf{r}/||\mathbf{r}||$ and a rotation angle $\Theta = ||\mathbf{r}||$. Orientation sampling is carried out by interpolation between two orientations $\mathbf{r_1}, \mathbf{r_2}$ using a spherical linear interpolation, i.e., Slerp \cite{shoemake1985}, which takes advantage of the fact that the spherical metric of $S^3$ is the same as the angular metric of $SO(3)$ for a constant angular speed of movement \cite{shoemake1985}. Moreover, it travels along the straightest (and shortest) path on the rounded surface of the quaternion unit sphere.

We first start by creating skew matrices $\mathbf{r_{+,1}}, \mathbf{r_{+,2}}$ from rotation vectors $\mathbf{r_1}, \mathbf{r_2}$.
These matrices can be easily converted to rotation matrices $\mathbf{R_1}, \mathbf{R_2}$ by computing their matrix exponentials (\textsl{expm} function):
$$ \mathbf{R_1} = \mathrm{expm}(\mathbf{r_{+,1}}), \qquad \mathbf{R_2} = \mathrm{expm}(\mathbf{r_{+,2}})$$
We compute the interpolation between rotations $\mathbf{R_1}, \mathbf{R_2} \in \mathbb{R}^{3\times 3}$ as
\begin{equation}
    \mathrm{slerp}(\mathbf{R_1}, \mathbf{R_2}, \alpha) = \mathrm{expm}\left( \alpha\,\mathrm{logm}\left(\mathbf{R_2R_1^T}\right) \right) \mathbf{R_1},
\end{equation}
where the function \textsl{logm} converts the rotation matrices to the skew matrices of the rotation vectors and $\alpha$ is an interpolation coefficient given as a period time $t_s$ divided by the trajectory time $T = \left(||\mathbf{x_d}-\mathbf{x_0}||/v_t\right)$ s.
Finally, the computed rotations are converted back to the axis-angle representation. 

\subsection{Peripersonal space projection}
\label{subsec:pps}
Roncone et al. \cite{roncone2016, roncone2015} proposed a distributed representation of the protective safety zone for humanoid robots, called Peripersonal space (PPS). The safety zone is represented by a collection of probabilities that an object from the environment eventually comes into contact with that particular skin taxel. For each taxel, they combined visual information of objects approaching the body and tactile information of eventual physical collision. A visual receptive field around the taxel is represented by a cone that rises from the surface along the normal. 

Nguyen et al. \cite{nguyen_merging_2018,nguyen_compact_2018} made small modifications to PPS and used it to ensure safe interaction between humans and robots. Instead of training the visual receptive fields, they defined them uniformly for all taxels. In addition, they extended the reach of the receptive field to a maximum of 45 cm. 
Due to compatibility with the original implementation, the taxel receptive field has a discrete representation with 20 bins representing the distance from the obstacle with values of collision probabilities. The Parzen window estimation algorithm interpolates the bins to create a continuous representation. A combination of the receptive fields constructs a safety volume margin around each body part.
In this work, we added receptive fields to the upper arm and torso to cover the entire upper body.

\subsection{Obstacle processing}
\label{subsec:obstacles}
The processed obstacles are the second input to the reactive controller (see the green box in Fig.~\ref{fig:overview}).
In our setup, we operate with obstacles detected by visual, proximity, and tactile sensors. 
However, our approach is universal in modalities. The processed obstacles, called collision points ($C$), are represented by their projected position on the robot $\mathbf{P}_C \in \mathbb{R}^{3}$, collision direction $\mathbf{n}_C \in \mathbb{R}^{3}$, threat level $a_{t}$ (i.e., scaled sensor measurement) and a gain $V_C$ representing the severity of the information; for example, tactile events can have the highest gain as the collision has already happened. The main advantage of our representation is that it jointly enables the post-collision reaction (tactile sensors) and collision avoidance (proximity and visual sensors).

An obstacle is incorporated into the QP problem as a linear inequality constraint that restricts movements towards the obstacle but allows motion in tangential directions. 
The constraint limits the approach motion of the collision point $C$ with Cartesian velocity $\mathbf{\dot{x}}_C \in \mathbb{R}^{3}$ towards the obstacle (in direction $\mathbf{n}_C$) by a maximum approach velocity $\dot{x}_a$ (which can be negative and become the repulsive velocity, Sec.~\ref{subsec:col_avoid_params}) as 
\begin{equation}
\mathbf{n}^T_C \mathbf{\dot{x}_C} = \mathbf{n}_C^T \mathbf{J}_C \mathbf{\dot{q}_C}  < \dot{x}_a
\end{equation}
\noindent
where $\mathbf{J}_C \in \mathbb{R}^{3\times n_c}$ is the translational-velocity component of the Jacobian associated with $C$ and $\mathbf{\dot{q}}_C \in \mathbb{R}^{n_c}$ are the joint velocities of the kinematic chain ending with $C$. 
The maximum approach velocity $\dot{x}_a$ is calculated from the obstacle threat level $a_{t}$ as
\begin{equation}
\label{eq:k_params}
\dot{x}_a = (k_1 - V_C \cdot a_t) \cdot k_2
\end{equation} 
where $k_1$ and $k_2$ are positive values to adjust the constraint which are described in Sec.~\ref{subsec:col_avoid_params}. 
In accordance with Eq. \ref{eq:prob_formulation}, we can write the obstacle constraints in matrix form ($\mathbf{A^i_o} \leq b^i_o$) as 
\begin{equation}
    \mathbf{n_C}^T \mathbf{J_C}  \leq (k_1 - V_C \cdot a_{t}) \cdot k_2.
\end{equation}

\begin{figure*}[htb]
    \centering
\begin{subfigure}[t]{0.28\textwidth}
    \centering
    \includegraphics[height=0.25\textheight]{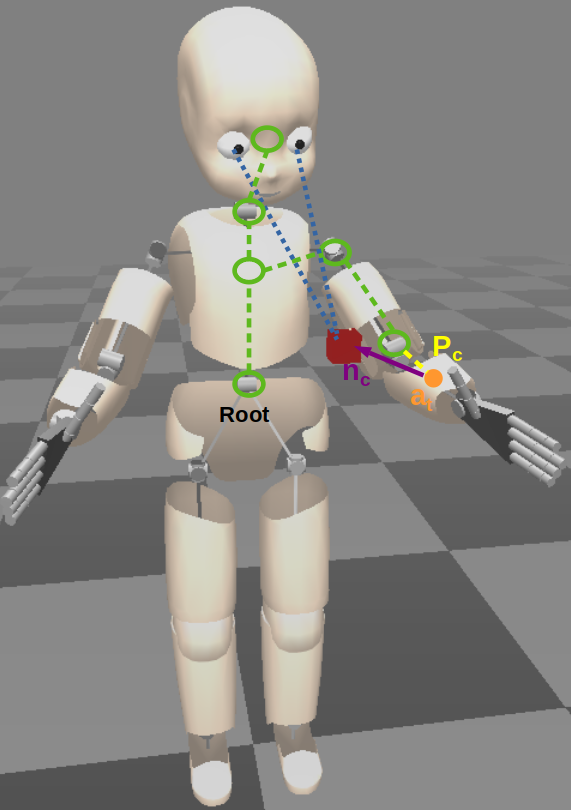}
    \caption{Visual obstacle projected onto the left forearm. The blue lines symbolize the lines of sight of the eyes, not those of the RGB-D camera for simplification.}
    \label{fig:visu-stimuli}
\end{subfigure}\hfil
\begin{subfigure}[t]{0.28\textwidth}
    \centering
    \includegraphics[height=0.25\textheight]{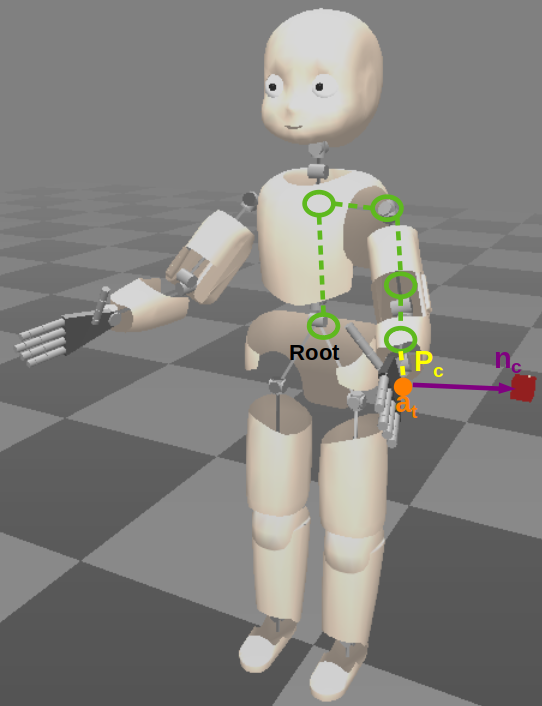}
    \caption{Proximity obstacle mapped to a sensor position on the left hand.}
    \label{fig:prox-stimuli}
\end{subfigure}\hfil
\begin{subfigure}[t]{0.28\textwidth}
    \centering
    \includegraphics[height=0.25\textheight]{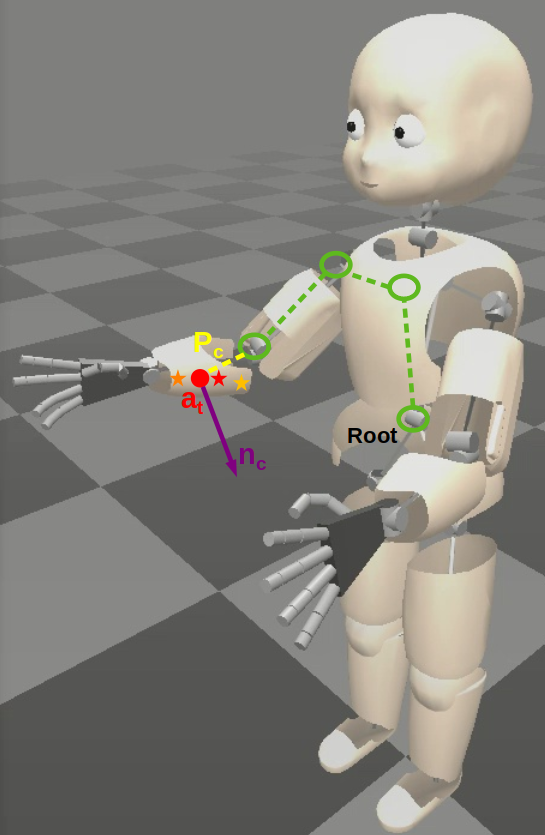}
    \caption{Several tactile stimuli on the right forearm (stars) combined into a super contact $P_C$ with a threat level of the highest measured pressure (red color).}
    \label{fig:tact-stimuli}
\end{subfigure}
\caption{Processing of obstacles. Obstacles (red cubes or stars) are projected onto the robot body ($P_C$) and connected (yellow line) to the robot kinematic chain (green line). Each collision point is characterized by a normal direction (purple arrow) and threat level (color of the circle, red means highest threat).}
    \label{fig:obstacle_scheme}
\end{figure*}

To prevent jerky motions induced by vanishing obstacle readings, we apply a linear decay formula to simulate an obstacle slowly moving away from the robot \cite{escobedo_contact_2021}. Every obstacle virtually exists for a specific amount of time (called ``surviving time''). For example, the threat disappears once the robot moves away from the tactile collision. Without virtual preservation of the obstacle, the robot would return to its original position even when the obstacle is still there.

Visual obstacles in our experiments are human keypoints detected by an RGB-D camera. The method for their detection is based on the work by Docekal et al. \cite{docekal2022}. The detected obstacles (see a schematic illustration in Fig.~\ref{fig:visu-stimuli}) are mapped to parts of the skin where a collision is most likely to occur using a PPS projection, similar to \cite{nguyen_compact_2018}. For each body part, we compute one visual obstacle as a weighted average of the mappings to that body part. The collision direction is the vector from the estimated contact position on the skin to the obstacle and the threat level is calculated from their distance.

Proximity sensor data (see a schematic illustration in Fig.~\ref{fig:prox-stimuli}) are not projected onto skin taxels, as these sensors are placed to cover parts of the robot without the skin; therefore, there is no skin taxel to which information can be projected. However, they are projected onto the sensor position. The threat level is computed from the measured distance, and the collision direction is the direction vector of the sensor beam. 

Tactile obstacles (see a schematic illustration in Fig.~\ref{fig:tact-stimuli}) are processed in the same way as in \cite{nguyen_merging_2018}. Contacts detected by artificial skin sensors (taxels) with a pressure above the threshold value are clustered to super contacts (one per each body part) to diminish spiking effects and eliminate false positive detections. The threat level corresponds to the highest measured pressure in the neighborhood, and the collision direction is the normal vector of the central taxel.

\subsection{Collision avoidance parameters}
\label{subsec:col_avoid_params}
Obstacle avoidance constraints (Eq.~\ref{eq:k_params}) are parameterized by two positive parameters---additive $k_1$ and multiplicative $k_2$. The parameter $k_1$ defines when the maximum approach velocity becomes the minimum repulsive velocity, that is, when $\dot{x}_a$ is lower than zero. The parameter $k_1$ has a constant value of 0.3 in our experiments.
The parameter $k_2$ scales the maximum approach velocity. We found that body links cannot move with the same minimum repulsive velocity due to the kinematic constitution of the robot. The same scale for all parts leads to unfeasible constraints for the upper arm and torso (high value) or small and slow avoidance for the hand and forearm (low value). 
From our observation, $k_2$ should increase from the torso to the hand. In our experiments, $k_2$ is 0.06, 0.06, 0.33, 0.53 for the body parts of the torso, upper arm, forearm, and hand, respectively.

\subsection{Self-collision avoidance}
\label{subsec:selfcol}
For self-collision avoidance, we sampled each body link to obtain sets of points on the robot surface. These points serve as both virtual obstacles and projected positions on the robot, which are the closest points of the robot to obstacles.
Due to the kinematic limitations of our humanoid robot, the projected positions are only on the hand and forearm of the controlled arm(s), and the virtual obstacles are on the common torso and the ``primary arm'' (hand, forearm, upper arm) when the dual-arm version of the controller is used. The virtual obstacles are not on the secondary arm because of the arm prioritization---only the ``secondary arm'' avoids the collision with the ``primary arm'' and not vice versa. 

We find the closest pair of virtual obstacles and projected positions for each combination of body parts. If their distance $d_C$ is less than a threshold, we add the virtual obstacle to the set of obstacles with $a_t \cdot V_C = 1.2 - 20d_C \geq 0$. The coefficients were empirically determined in simulation and on the real robot to obtain appropriate self-collision avoidance reactions.
The collision point is the projected position of the obstacle on the robot ($\mathbf{P}_C$), the collision direction ($\mathbf{n}_C$) is the vector from the collision point to the virtual obstacle.
Finally, we compute the linear inequality constraint for these obstacles in the same way as for the external obstacles.

\subsection{Static obstacles}
\label{subsec:static_obs}
Sometimes, the experiment contains known static objects that the robot should avoid. The almost same procedure as for self-collision avoidance can also be used in this case. Virtual obstacles are sampled from the static object instead of the robot body. The projected positions are points on the robot body parts that are in danger of colliding with the object. 
This approach was used, for example, to avoid the table during the interactive game demonstration. 

\subsection{Relative position constraint}
\label{subsec:relpos_constr}
Other task restrictions can be incorporated into the problem as (in-)equality constraints. For the bimanual task experiment, we added one constraint per each axis to keep a specified relative position between the hands (end effectors).
The constraint can be written in the vector form as
$$ \mathbf{x^p_{k+1}} - \mathbf{x^s_{k+1}} = \mathbf{d_{\mathrm{rel}}} $$
where $\mathbf{x^{p/s}_{k+1}} \in \mathbb{R}^{3}$ are the next (i.e. in timestep $k+1$) positions of the end effectors, and $\mathbf{d_{\mathrm{rel}}} \in \mathbb{R}^{3}$ is a relative position vector. 
This equation can be reformulated for our QP problem with joint velocities as 
\begin{equation}
\begin{split}
\mathbf{x_{k}^p} + \begin{bmatrix}\mathbf{J_{t,pos}^p}(\mathbf{q^t}) & \mathbf{J_{a,pos}^p}(\mathbf{q^p})\end{bmatrix}
\begin{bmatrix}\mathbf{\dot{q}^t} \\ \mathbf{\dot{q}^p}\end{bmatrix}& - \\ \mathbf{x_{k}^s} -\begin{bmatrix}\mathbf{J_{t,pos}^s}(\mathbf{q^t}) & \mathbf{J_{a,pos}^s}(\mathbf{q^s})\end{bmatrix}
\begin{bmatrix}\mathbf{\dot{q}^t} \\ \mathbf{\dot{q}^s}\end{bmatrix}& = \mathbf{d_{\mathrm{rel}}}
\end{split}
\end{equation}
\noindent
where $\mathbf{x_{k}^{p/s}} \in \mathbb{R}^{3}$ are the current (i.e. in timestep $k$) positions of the end effectors, $\mathbf{J^{p/s}_{a,pos}}(\mathbf{q^{p/s}}) \in \mathbb{R}^{3\times (n-3)}, \mathbf{J^{p/s}_{t,pos}}(\mathbf{q^t})\in \mathbb{R}^{3\times 3}$ are the translational velocity components of the Jacobian matrices for arm and torso, respectively.

\subsection{Joint position constraints}
\label{subsec:joint_constr}
The joint position bounds for the QP problem must be set in such a way that the robot joint limits are not exceeded. In our solution, we are inspired by the shaping policy to avoid joint limits in \cite{pattacini_experimental_2010} to maximize reachability in the arm workspace.
It consists of a flat region in most of the joint range. We replaced the original hyperbolic tangent functions with linear and constant functions to fulfill the constraint condition of the QP problem.
Since the problem is solved at the velocity level, the joint position limits are expressed using the joint velocities for the joint $i$ as:
$$\dot{q}^i_l = \dot{q}^i_{\mathrm{min}}\cdot c^i_{\mathrm{min}},\qquad \dot{q}^i_u = \dot{q}^i_{\mathrm{max}}\cdot c^i_{\mathrm{max}}$$
\noindent
where $\dot{q}^i_{\mathrm{min}}$ and $\dot{q}^i_{\mathrm{max}}$ are minimum and maximum permissible joint velocity respectively, $c^i_{\mathrm{max}}$ and $c^i_{\mathrm{min}}$ are obtained from the joint limit avoidance policy as:
\begin{equation}
\begin{split}
c^i_{\mathrm{min}} &= \begin{cases}
0 & q^i < g_L^i\\
\frac{q^i - g_L^i}{g_H^i - g_L^i} & g_L^i <= q^i <= g_H^i\\
1 & g_H^i < q^i\end{cases} \\
c^i_{\mathrm{max}} &= \begin{cases}
0 & G_H^i < q^i\\
\frac{q^i - G_H^i}{G_L^i - G_H^i} & G_L^i <= q^i <= G_H^i\\
1 & q^i < G_L^i\end{cases}
\end{split}
\end{equation}
where $q_i$ is a current joint position, $g_L^i$ and $G_H^i$ are the lowest and highest permissible joint position value respectively, and $g_H^i$ and $G_L^i$ are the thresholds for the joint position constraints.

\section{Experiments and Results}
We prepared seven experiments with a simulated iCub robot and three with a real one to validate the performance of the proposed solution and compare it with other solutions. An overview of the experiments can be seen in Tab.~\ref{tab:experiments}.

\begin{table}[htb]
\centering
\resizebox{245pt}{!}{%
\begin{tabular}{c|ccccc}
\textbf{Exp} & \textbf{Type} & \textbf{Movement} & \textbf{Purpose} & \textbf{Targets} & \textbf{Obstacles} \\ \hline
1 & Sim & P2P & Reachability & 3x3x3 grid & - \\ \hline
2 & Sim & P2P & Smoothness & 2 poses & - \\\hline
3 & Sim & Circular & \begin{tabular}[c]{@{}c@{}}Reference \\ tracking\end{tabular} & r = 0.08 m & - \\ \hline
4 & Sim & Circular & \begin{tabular}[c]{@{}c@{}}Self-Coll. \\ avoidance\end{tabular} & r = 0.1 m & - \\ \hline
5-1 & Sim & P2P & \begin{tabular}[c]{@{}c@{}}Collision\\ avoidance\end{tabular} & 2 poses & \begin{tabular}[c]{@{}c@{}}1 with vel. \\ {[}0, -0.05, 0{]}\end{tabular} \\ \hline
5-2 & Sim & P2P & \begin{tabular}[c]{@{}c@{}}Collision\\ avoidance\end{tabular} & 2 poses & \begin{tabular}[c]{@{}c@{}}1 with vel. \\ {[}0.05, 0, 0{]}\end{tabular} \\ \hline
5-3 & Sim & P2P & \begin{tabular}[c]{@{}c@{}}Collision\\ avoidance\end{tabular} & 2 poses & \begin{tabular}[c]{@{}c@{}}1 with vel. \\ {[}0, 0, -0.05{]}\end{tabular} \\ \hline
11 & Real & P2P & \begin{tabular}[c]{@{}c@{}}Obstacle\\ management\end{tabular} & 1 pose & \begin{tabular}[c]{@{}c@{}}multiple\\ obstacles\end{tabular} \\ \hline
12 & Real & P2P & \begin{tabular}[c]{@{}c@{}}Obstacle\\ management\end{tabular} & 2 poses & \begin{tabular}[c]{@{}c@{}}multiple\\ obstacles\end{tabular} \\ \hline
13 & Real & Bimanual & \begin{tabular}[c]{@{}c@{}}Whole\\ solution\end{tabular} & 1 pose & \begin{tabular}[c]{@{}c@{}}multiple\\ obstacles\end{tabular}
\end{tabular}
}
\caption{Experiments overview; P2P stands for point-to-point movement.\label{tab:experiments}}
\end{table}

\subsection{Experiments in simulation}
\label{subsec:sim_results}
In simulation, we evaluated the performance of the controller (Sec.~\ref{subsec:controller}) and the trajectory sampling (Sec.~\ref{subsec:local_sampling}). 
The first set of experiments aims to assess reachability, movement smoothness, and reference tracking, without obstacles nearby.
Reachability is tested in a reaching experiment (Exp 1) with the targets taken from a grid of 3x3x3 positions in Cartesian space in random order and randomly switched two orientations. The positions are the Cartesian product of $x =$ [-0.23, -0.19, -0.15] m, $y =$ [0.11, 0.15, 0.19] m, $z =$ [0.08, 0.12, 0.16] m, and the orientations are $o_1 =$ [-0.15, -0.79, 0.59, 3.06], $o_2 =$ [-0.11, 0.99, 0.02, 3.14] in angle-axis notation.
Smooth movement is evaluated in a point-to-point experiment (Exp 2), where the robot moves between two alternating poses $p_1 =$ [-0.23, 0.26, 0.02, -0.15, -0.79, 0.59, 3.06] and $p_2 =$ [-0.26, 0.03, 0.03, -0.11, 0.99, 0.01, 3.14], where the first three values are position coordinates in meters and the last four values are the orientation in angle-axis notation.
Finally, a circular movement experiment (Exp 3) is chosen to validate the reference tracking. The overview of the experiments can be found in Tab.~\ref{tab:experiments}.

In addition to the proposed solution, other alternatives and state-of-the-art methods are also evaluated. 
Specifically, we compare --- 1) \textsl{HARMONIOUS} -- the proposed solution in the single-arm version; 2) \textsl{HARMONIOUS2} -- the proposed solution in the dual-arm version when the goal of the other arm is to hold a specified position; 3) \textsl{NEO} -- the controller by Haviland and Corke \cite{haviland_neo_2021}; 4) \textsl{React} by Ngyuen et al.~\cite{nguyen_merging_2018}; 5) \textsl{Cart} -- Cartesian controller for the iCub robot \cite{pattacini_experimental_2010}, which is evaluated only in Exp 1 and Exp 2 as it primarily serves for point-to-point movements and does not feature avoidance constraints. Controllers 1--4 are evaluated with and without local trajectory sampling (LTS), described in Sec.~\ref{subsec:local_sampling}. \textsl{Cart} has its own local trajectory optimization.

\begin{table}[tb]
\centering
\begin{tabular}{c|ccccc}
LTS & HARMON & HARMON2 & NEO & React & Cart \\ \hline
Yes & 91.1 \% & 90.4 \% & 77.8 \% & 86.7 \% & \multirow{2}{*}{90.4 \%} \\
No & 90.4 \% & 89.6 \% & 77.8 \% & 82.2 \% & 
\end{tabular}
\caption{Reachability experiment (Exp 1). Success rate of combined reachability in position and orientation. LTS stands for local trajectory sampling. \textsl{Cart} has its own local trajectory optimization. The success rate is computed from 5 repetitions (with a different order of the targets each time) of the experiment (in total 135 targets).}
\label{tab:reach}
\end{table}

The results of the reachability experiment (Exp 1) are shown in Tab.~\ref{tab:reach}. In this experiment, our controller has the highest number of targets that are reached in 6D pose (i.e., position and orientation). The target is reached when the end effector is, at the same time, closer than 5 mm and 0.1 rad to the target. The success rate is computed from 5 repetitions (with a different order of the targets each time) of the experiment (in total 135 targets).
The impact of the local trajectory sampling is visible as all controllers have better (or at least the same) reachability results with sampling than without sampling.

\begin{figure*}[htbp]
    \centering
\begin{subfigure}[t]{0.65\textwidth}
    \centering
    \includegraphics[width=\textwidth]{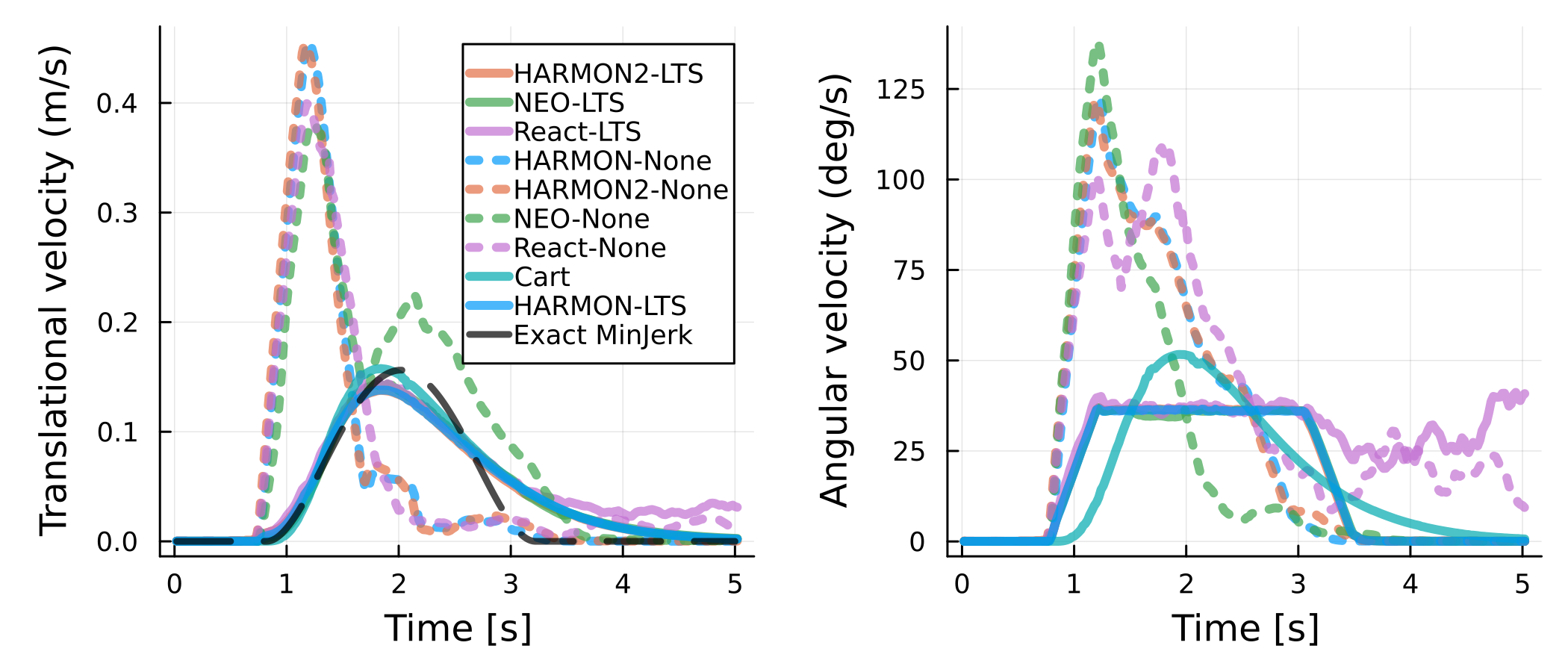}
    \caption{Exp 2: End-effector translational (left) and angular (right) velocity profiles during point-to-point movement and computed exact minimum-jerk translational velocity profile.}
    \label{fig:exp2}
\end{subfigure}\hfill
\begin{subfigure}[t]{0.3\textwidth}
    \centering
    \includegraphics[width=0.9\textwidth]{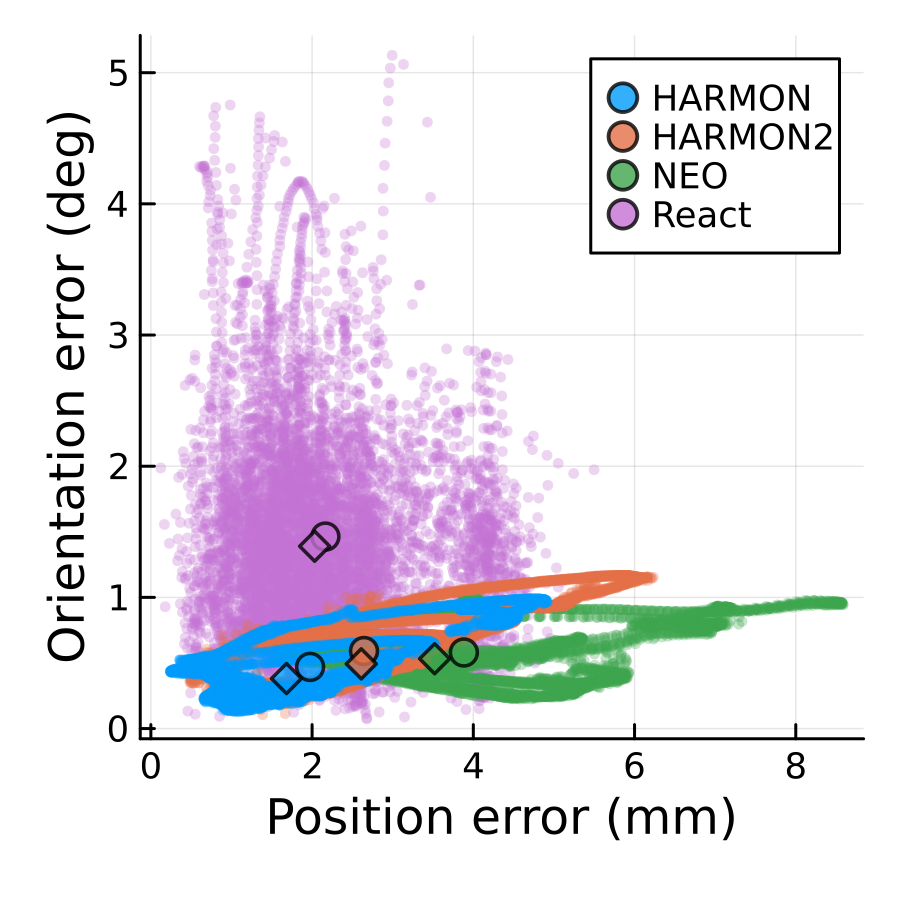}
    \caption{Exp 3: Reference tracking error between desired and actual 6D pose. Circles and diamonds represent mean and median pose error, respectively.}
    \label{fig:exp3}
\end{subfigure}

\begin{subfigure}[b]{0.39\textwidth}
    \centering
    \includegraphics[width=\textwidth]{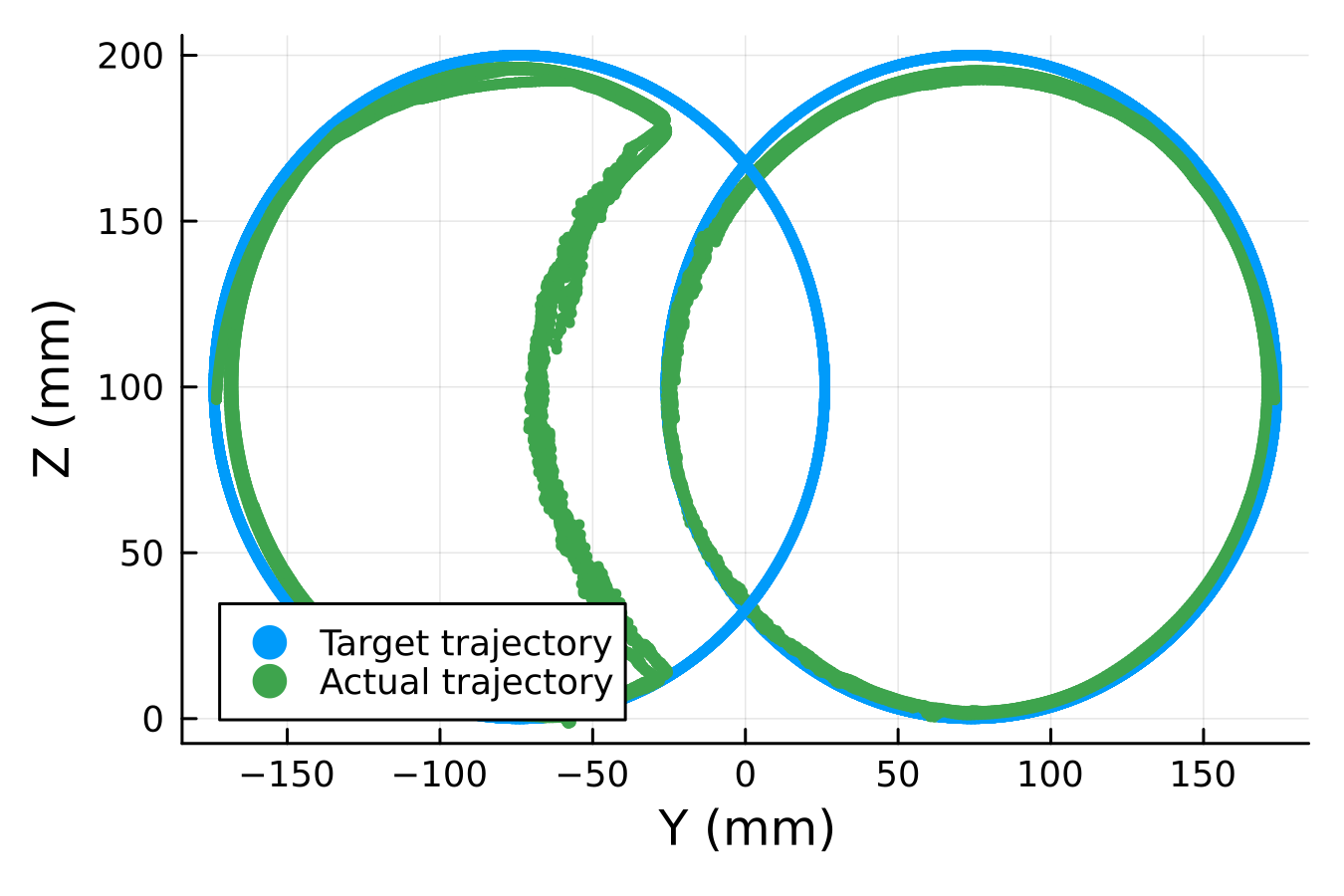}
    \caption{Exp 4: Left and right arms movement for our solution (right set as ``primary'').}
    \label{fig:exp4}
\end{subfigure}\hfill
\begin{subfigure}[b]{0.6\textwidth}
    \centering
    \includegraphics[width=\textwidth]{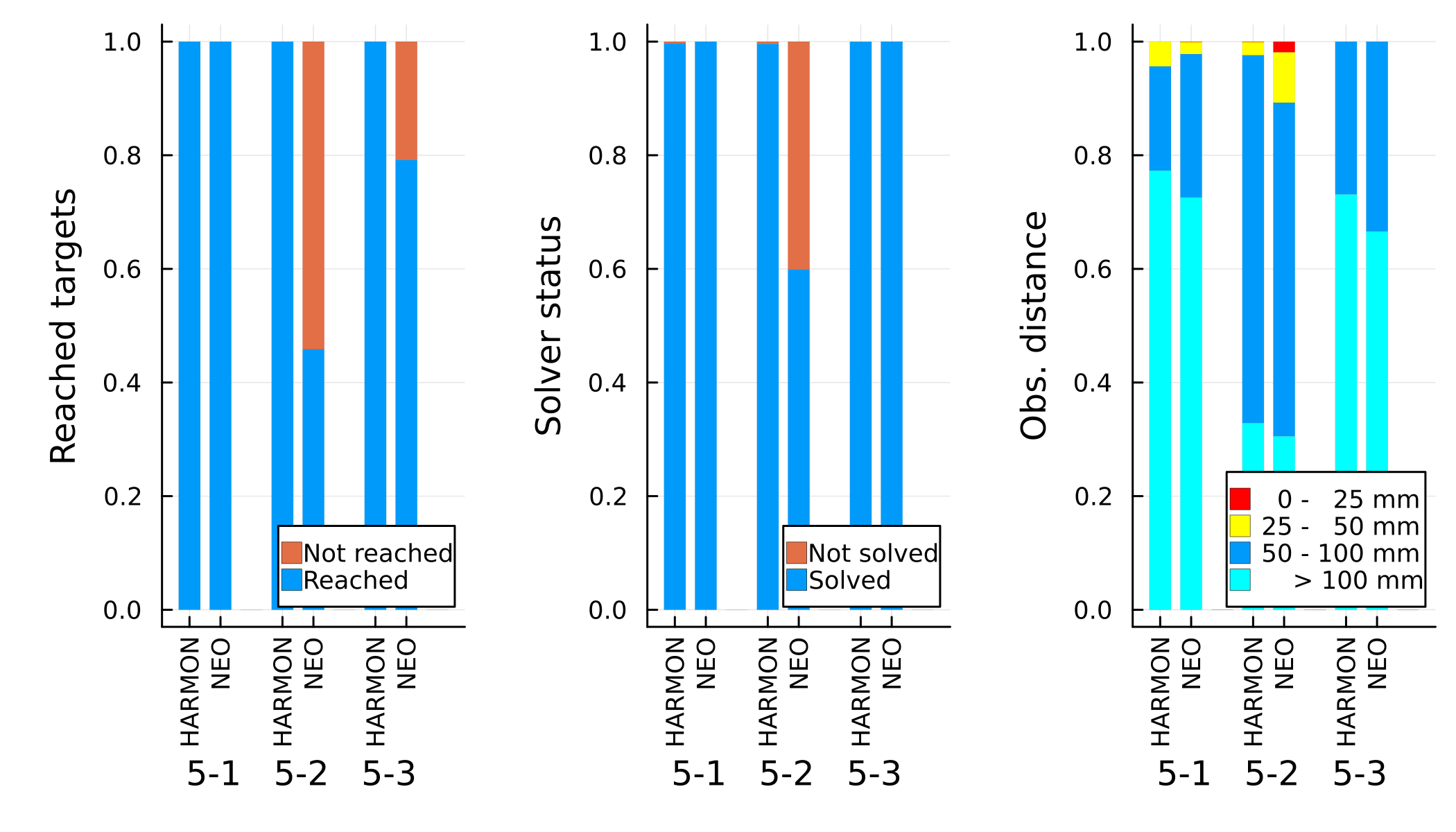}
    \caption{Exp 5: Collision avoidance experiments (Exp 5-1, 5-2, 5-3).}
    \label{fig:exp5}
\end{subfigure}
    \caption{Simulation experiments results. LTS stands for local trajectory sampling.}
    \label{fig:results_sim}

\end{figure*}

The results of the movement smoothness experiment (Exp 2) show the main impact of the local trajectory sampling (see  Fig.~\ref{fig:exp2}). Position sampling (Fig.~\ref{fig:exp2} left) creates a bell-shaped Cartesian translation velocity profile typical of human arm motion \cite{Flash1985}. The velocity profile of \textsl{Cart} is the most similar to the exact minimum-jerk movement, as it is specially optimized for it. Other versions are almost identical; only \textsl{React} does not reach zero velocity after 5~s. All versions without local sampling have snap onset and slow jerky decay.
Orientation sampling (Fig.~\ref{fig:exp2} right) creates a torque-minimal path for rotation as it travels along the straightest path of the rounded surface of a sphere and keeps the rotational velocity constant during movement \cite{shoemake1985}, resulting in a visually smooth transition between two rotations. \textsl{HARMONIOUS} and \textsl{NEO} have a smooth angular velocity profile; \textsl{React} velocity profile is not smooth and does not reach zero after 5~s. \textsl{Cart} uses the same sampling for orientation as for position, and thus the resulting angular velocity profile is similar to that for translational velocity. 

In the case of circular movement in the reference tracking experiment (Exp 3), there is no effect of local trajectory sampling, as a new target is generated for each period (the tracked reference), meaning that the smoothness of the movement is handled at a higher level (an external target generator). As shown in Fig.~\ref{fig:exp3}, \textsl{HARMONIOUS} outperformed others in reference tracking with the lowest position and orientation error. 

The next experiment is a circular movement experiment with both arms in opposite directions (Exp 4). It shows how the dual-arm solution works, particularly arm prioritization and self-collision avoidance. As the reference controllers are for single arms only (or separately), only our solution is shown (without trajectory sampling). Figure~\ref{fig:exp4} shows the trajectories of both end effectors making circular movements. In this case, the right arm was selected as the ``primary arm''. Therefore, the right arm successfully tracks the circle all the time and the left arm prevents self-collision by keeping a safe distance instead of tracking the circle when the arms are too close.

The last set of simulation experiments tests collision avoidance. For simplicity, the obstacle is virtual and is perceived without noise, omitting the influence of sensor data processing on the results. 
All three experiments are based on Exp 2 and differ only in the direction of the obstacle: 1) the obstacle moves along the movement, i.e., in the y-axis direction (Exp 5-1); 2) the obstacle moves towards the robot, i.e., in the x-axis direction (Exp 5-2); 3) the obstacle falls in the z-axis direction (Exp 5-3). 
In this case, \textsl{Cart} is not included in the comparison because it cannot handle obstacles. \textsl{React} is also omitted because it was outperformed already in the experiments without obstacles. More restrictive repulsive vectors were used instead of linear inequality constraints that allow motions in tangential directions. Therefore, we compare only \textsl{HARMONIOUS} and \textsl{NEO}. 
Figure~\ref{fig:exp5} shows the fraction of targets reached during the experiment, the fraction of QP problems successfully solved during the experiments, and the distance between the obstacle and the closest point of the robot, where we assume that a distance of less than 25 mm is too dangerous.
There is no significant difference between controllers when the obstacle moves along the robot's movement (Exp 5-1). On the other hand, the obstacle that moves towards the robot (Exp 5-2) is problematic for \textsl{NEO}. Half of the time, the targets are not reached within a time limit and the controller cannot solve the QP problem in almost 40 \% runs. Furthermore, small distances between the robot and the obstacle sometimes lead to collisions.
Unlike \textsl{NEO}, \textsl{HARMONIOUS} has no problem with collisions or reaching in this experiment (see video S1 in Multimedia Materials). Our solution also works adequately in Exp 5-3. \textsl{NEO} did not reach all targets within the time limit, but the distance to an obstacle is safe.

\subsection{Real robot experiments -- collision avoidance}
\label{subsec:results_real}
The experiments with the real robot were prepared to evaluate the obstacle processing part of the solution, described in Sec.~\ref{subsec:obstacles}. We show collision avoidance (visual and proximity modalities) and post-collision reaction (tactile modality) and their combination while the robot keeps position (Exp 11, see video S2 in Multimedia Materials) or moves back and forth between two points as in Exp 2 (Exp 12, see video S3 in Multimedia Materials). 
When visual obstacles are involved, the iCub gaze controller \cite{Roncone2016RSS} is used to point the robot gaze to an appropriate position in the scene, thus compensating for the robot torso joint movements.

\begin{figure}%
  \centering
  \begin{subfigure}[t]{0.489\textwidth}
  \centering
    \begin{minipage}[c]{0.5\textwidth}
    \includegraphics[width=\textwidth]{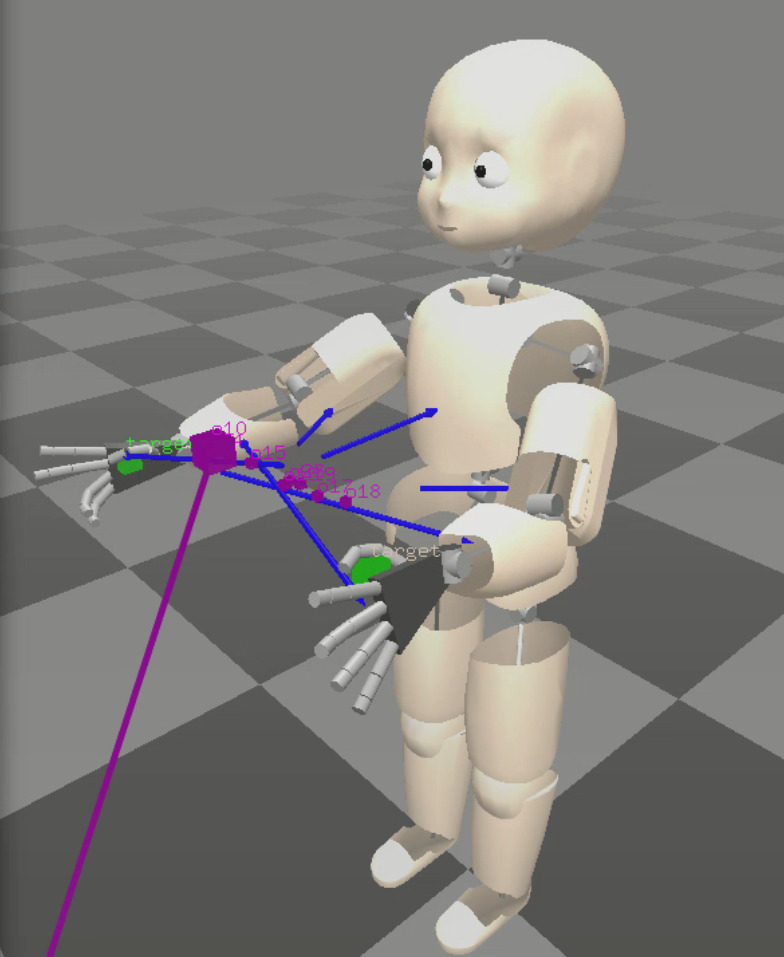} 
  \end{minipage}
  \begin{minipage}[c]{0.32\textwidth}\caption{Visualization of visual obstacle avoidance. Green cubes -- target positions for the robot end effectors; purple cubes -- human keypoints detected, larger cubes for body and smaller for hand keypoints; blue arrows -- PPS projections of obstacles onto robot body parts.\label{fig:holdpos_gui}}
  \end{minipage}
  \end{subfigure}\vspace*{2mm}
    \subcaptionbox{RGB-D camera view of a human approaching the robot with keypoint detections.}{\includegraphics[width=0.35\textwidth]{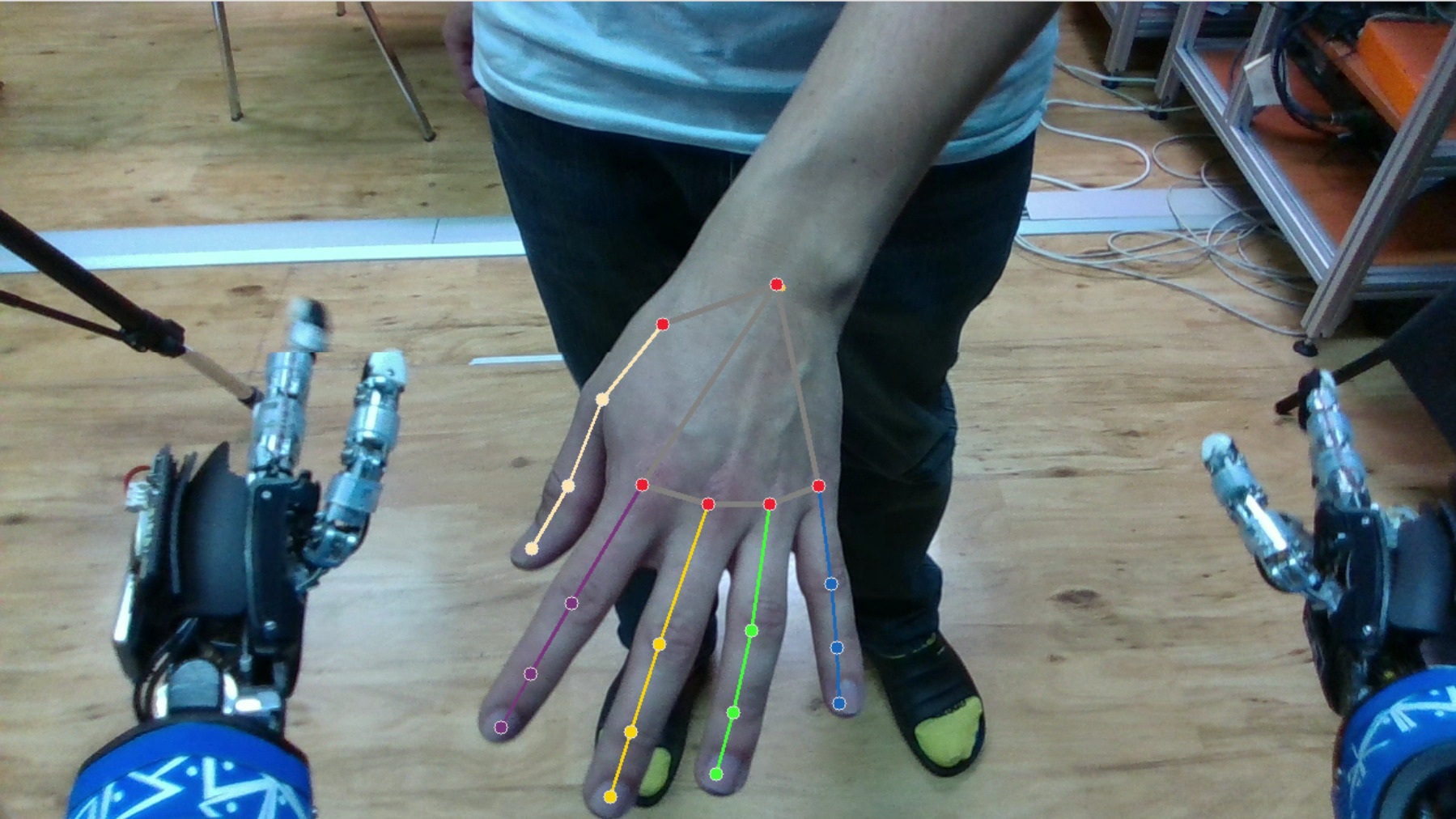} \label{fig:holdpos_rgbd}}
    \subcaptionbox{External view -- human approaching the robot.}{\includegraphics[width=0.35\textwidth]{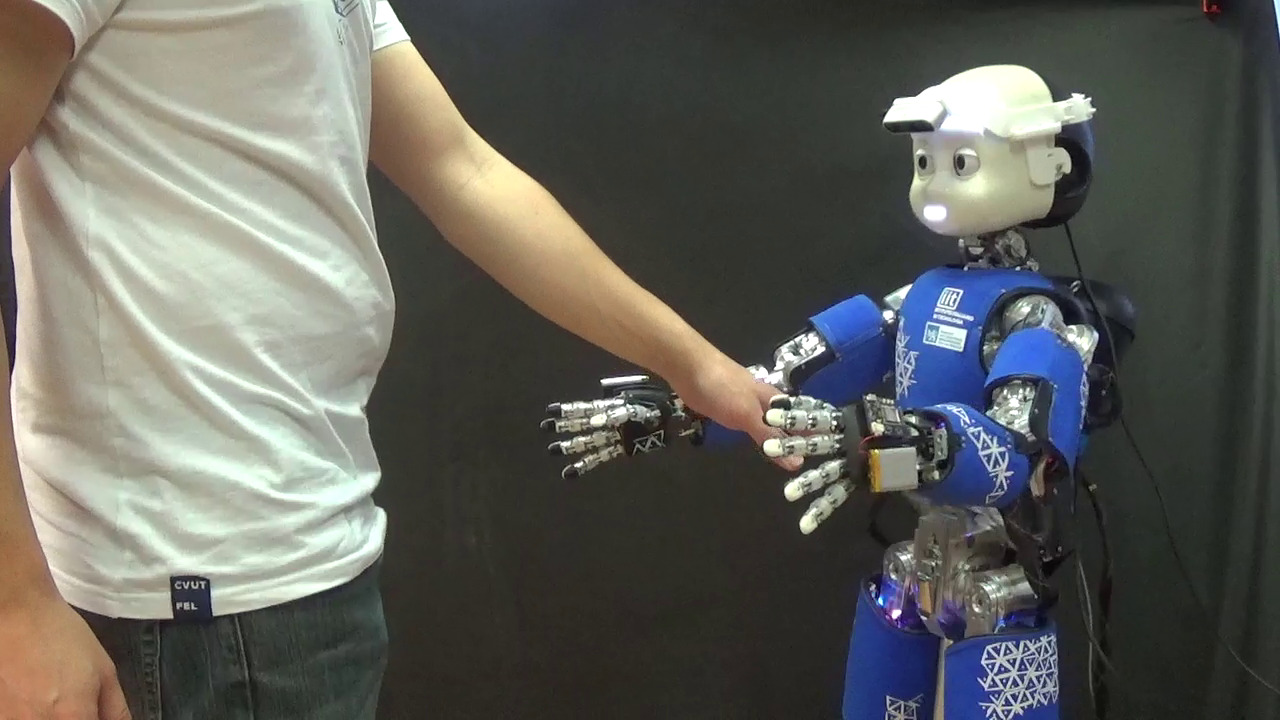} \label{fig:holdpos_ext}}%
  \caption{Visualizations from real-world collision avoidance experiments. \label{fig:exp112_photos}}%
\end{figure}

\begin{figure}[htbp]
    \centering
    \begin{subfigure}[t]{0.24\textwidth}
    \centering
    \includegraphics[width=\textwidth]{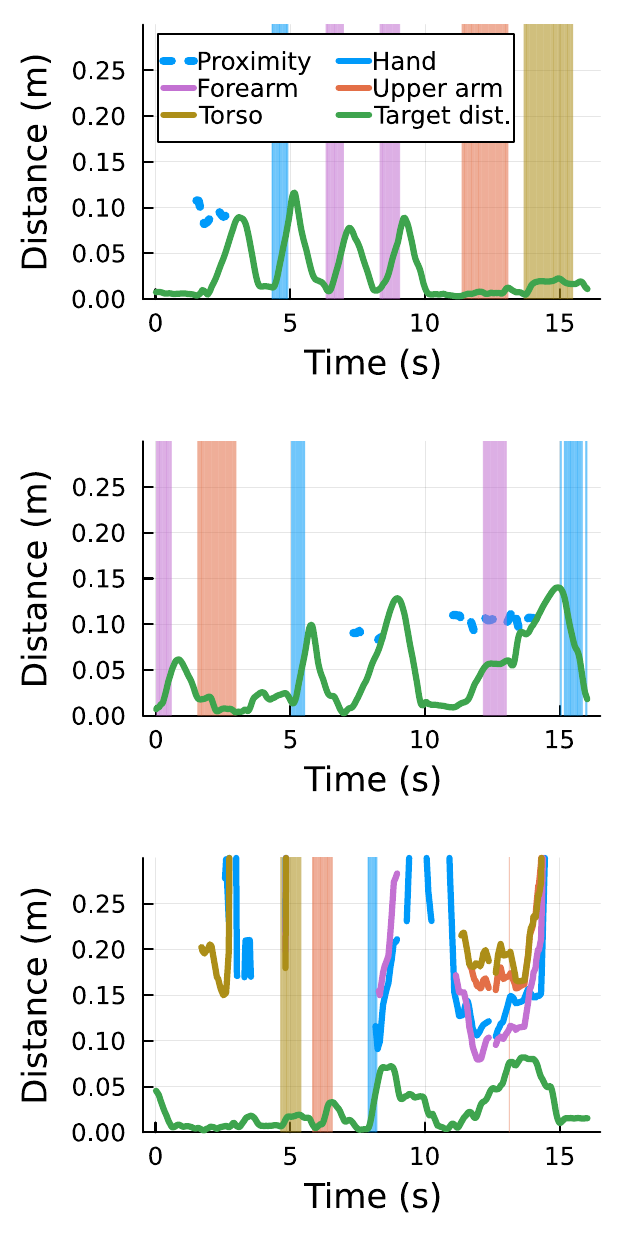}
    \caption{Exp. 11 -- Left arm keeping position. Tactile and proximity collision avoidance (top and center), tactile and visual collision avoidance (bottom).\label{fig:exp11_res}}    
    \end{subfigure}
    \begin{subfigure}[t]{0.24\textwidth}
    \centering
    \includegraphics[width=\textwidth]{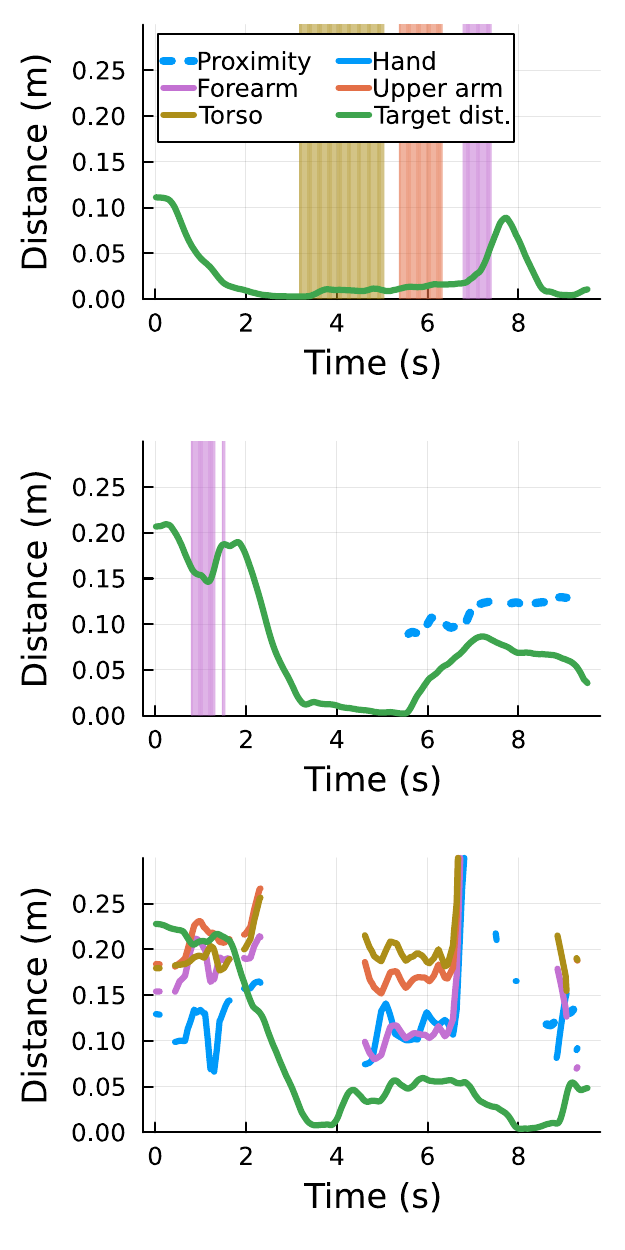}
    \caption{Exp. 12 -- Left arm reaching. Tactile collision avoidance (top), proximity and tactile collision avoidance (center), visual collision avoidance (bottom).\label{fig:exp12_res}}    
    \end{subfigure}
    \caption{Real-world collision avoidance experiments. Green line – distance of end effector (robot hand) to target; solid lines in other colors – distance of visual obstacles to the closest body part; dotted line –- distance to back of robot hand from proximity sensor; vertical lines/stripes –- tactile activations -– contacts with corresponding body parts.}
    \label{fig:exp112_results}
\end{figure}

Figure~\ref{fig:exp112_photos} graphically illustrates the experiment using snapshots. Figure~\ref{fig:exp112_results} shows the operation of the main components of \textsl{HARMONIOUS} on plots, focusing on the left arm of the robot (the ``primary arm'' in these experiments). 
Note that obstacles at or near the robot hand (visual obstacles near the hand or forearm, proximity at the back of the hand, or touch at the hand or forearm) inevitably compromise the task---staying close to the target---whereas visual obstacles or touch at more proximal body parts give the robot the possibility to exploit its kinematic redundancy and fulfill the task while simultaneously avoiding contacts. 
Parts of Exp 11---robot tasked with keeping the end effector position while complying with constraints from obstacles spawned by the presence of the human---are shown in Fig.~\ref{fig:exp11_res}. In Fig.~\ref{fig:exp11_res} (top), the robot starts in a steady state without obstacles around characterized by a zero target distance (first two seconds, green line). Once a proximity obstacle is detected, the end effector is moving away until the proximity obstacle constraint disappears from the controller and the target position is reached again. Physical contact of the human with the robot hand (around 5~s) and the forearm (around 7 and 9~s) generate constraints and trigger avoidance. Tactile collision with upper arm (12-13~s) and torso (14-15~s) generate avoidance but do not significantly compromise the task, as the robot kinematic redundancy is exploited by the controller. 

In Fig.~\ref{fig:exp11_res} (center), the first 10 seconds are very similar to the top one. Touch on the forearm causes deviation from the target while contact with the upper arm is evaded while simultaneously staying close to the target. Touch at the hand (5~s) and proximity signal at the hand (7~s) cause deviation from the target. At around 12~s, a proximity obstacle at the hand is evaded, but then avoidance stops as the robot is blocked by contact with the forearm. Once the touch disappears, the proximity collision avoidance continues. Finally, when the proximity obstacle disappears, a tactile stimulus on the hand causes a fast movement towards the target position.

Figure~\ref{fig:exp11_res} (bottom) shows a part of the experiment where, in addition to proximity and tactile obstacles, visual obstacles are spawned by human keypoint detections in the camera input. At first, distant visual stimuli for the torso and hand cause only small end-effector deviations from the target. Then, at 5~s, a physical collision with the torso and upper arm occurred, outside the field of view. In contrast, the subsequent tactile input at the hand (8~s) is also detected by vision. Visual stimuli persist once the collision ends, preventing the robot from reaching the target until they disappear. At the end (12-15~s), we can see the Peripersonal space (PPS) projection to all parts of the skin causing deviation from the target. 

In Exp 12, the robot is asked to move between two targets every 9~s. Figure~\ref{fig:exp12_res} show the main system components in action. Figure~\ref{fig:exp12_res} (top) starts with an undisturbed reaching movement to the target, followed by touch on the torso (4~s) and upper arm (6~s) causing only small deviations from the target. To evade contact with the forearm (7~s), distance to target has to increase. In Fig.~\ref{fig:exp12_res} (center), the human physically interferes with the forearm during the approach phase, so the arm has to go around the obstacle to reach the target (2-4~s). After the target is reached, a proximity obstacle pushes the hand away from the target again (6-9~s). The effect of visually detected obstacles is shown in Fig.~\ref{fig:exp12_res} (bottom). At first, visual stimuli block the direct way to the target; thus, the arm goes around the obstacles to prevent collisions (0-4~s). Once the target is reached, new visual obstacles appear (5-7~s), forcing the end effector deviate from the target again. Note that no particular evasive maneuvers to go around an obstacle to the target are built in. Such behaviors are rather emergent from the local behavior of the controller. 

\subsection{Real robot -- bimanual task}
\label{subsec:bimanual}
\begin{figure}[htbp]
    \centering
    \begin{subfigure}[t]{0.235\textwidth}
    \includegraphics[width=\textwidth]{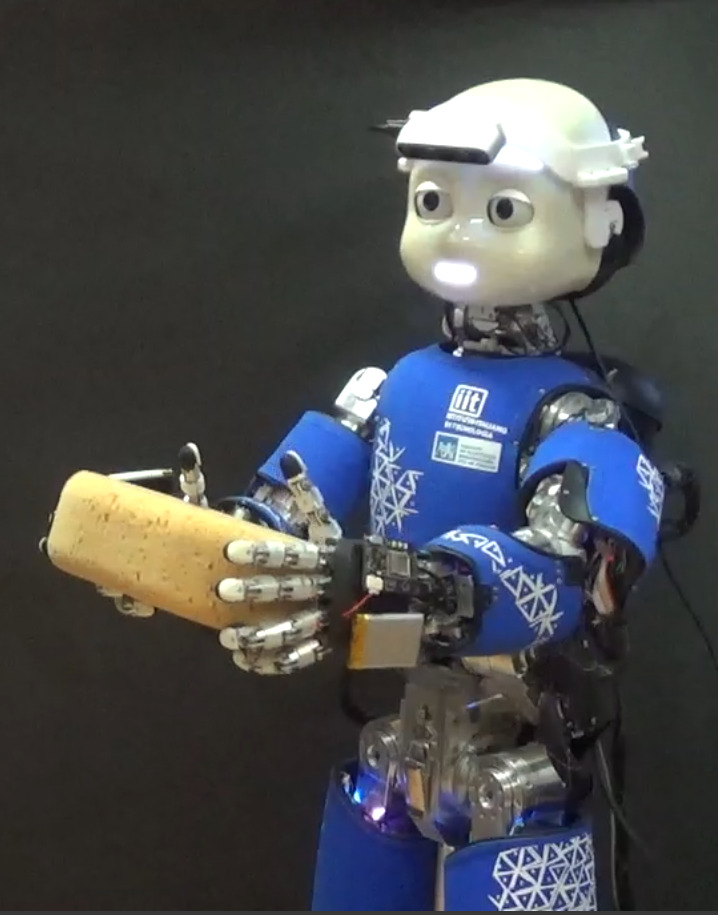}    
    \caption{Photo from the experiment.\label{fig:exp13_photo}}
    \end{subfigure}
    \begin{subfigure}[t]{0.245\textwidth}
        \includegraphics[width=\textwidth]{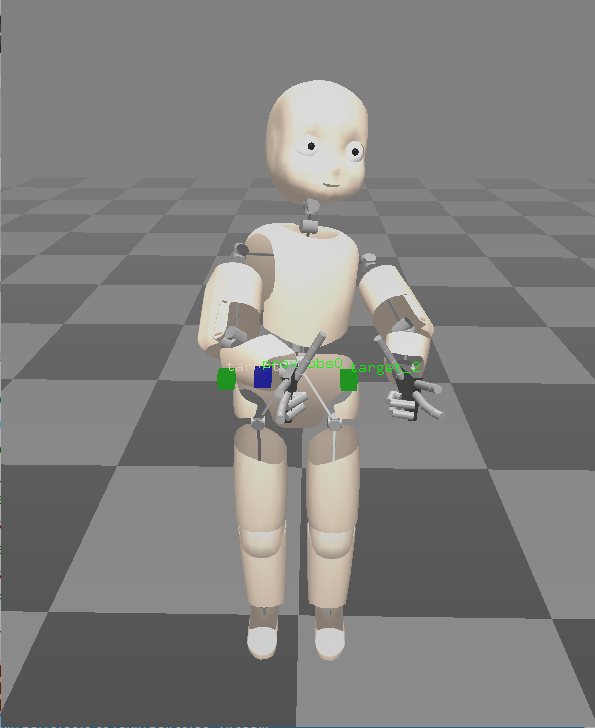}
        \caption{Visualization of hand targets (green cubes) and proximity obstacle (blue cube).\label{fig:exp13_visu}}
    \end{subfigure}

    \begin{subfigure}[t]{0.48\textwidth}
    \centering
        \includegraphics[width=0.92\textwidth]{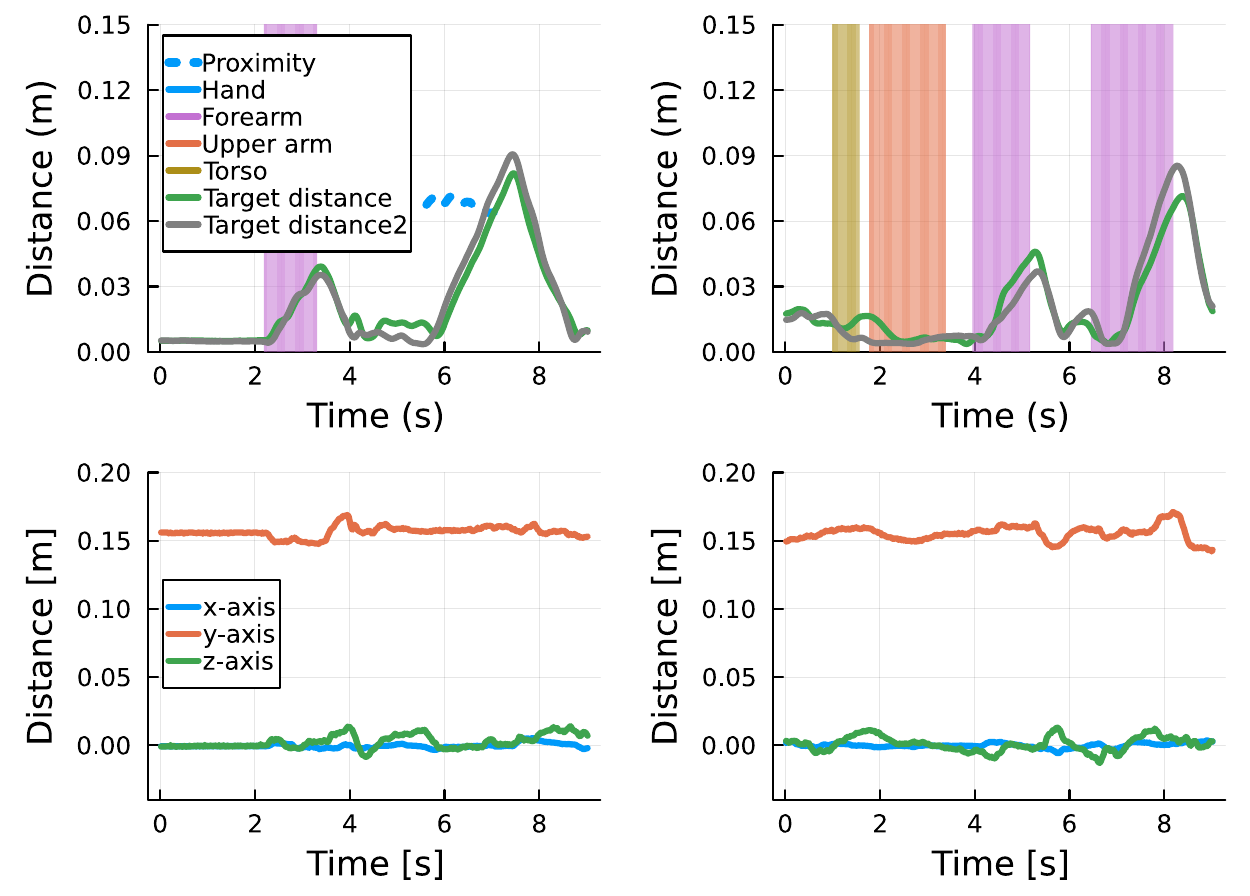}
        \caption{Two sequences from the bimanual task experiment in columns. Obstacle and target distances (top): green and gray lines -- distance of end effectors (robot hands) to target; dotted line -- distance to back of robot hand from proximity sensor;  vertical lines/stripes -- tactile activations -- contacts with corresponding body parts. Relative position of the end effectors (bottom).\label{fig:exp13_res}}
    \end{subfigure}
    \caption{Real robot -- bimanual task.}
    \label{fig:results_bimanual}
\end{figure}

We prepared a bimanual task experiment (Exp 13, see video S4 in Multimedia Materials) in which the robot holds a sponge in its hands (see Fig.~\ref{fig:exp13_photo}). The target is prescribed for the ``primary arm'' (left) and an additional constraint is introduced to keep the sponge between the two hands. At the same time, dynamic obstacles spawn additional constraints and avoidance. 
For this experiment, tactile inputs from the robot hands (palms) are ignored (the robot holds the sponge). This experiment aims to verify that the dual-arm version works and that other task constraints, such as the relative distance vector between the end effectors in this case, can be easily incorporated into the problem definition (the relative position constraint described in Sec.~\ref{subsec:relpos_constr}).

Figure~\ref{fig:exp13_res} shows plots from two sequences of the bimanual task experiment---distances to the target, presence of obstacles for the right and left arms, and the relative position of the end effectors. In the left column, we can see the evasive action of the arms caused by contact with the left forearm and then a proximity obstacle impinging on the left hand. The relative position between both hands remained almost the same as the arms simultaneously moved away from the obstacles, respecting the bimanual constraint. 
In the right column, we can again see how the high number of degrees of freedom helps to keep the target positions while reacting to contacts on the proximal body parts---torso and upper arm. Contact with the left forearm causes deviation from the target but not losing or squishing the sponge.

Collision avoidance during the bimanual task is more challenging, as it requires a synchronized movement of both arms to keep the specified relative position between them. Therefore, the avoidance movements are smaller compared to previous experiments. The proximity obstacle avoidance is portrayed in Fig.~\ref{fig:exp13_visu}.

\subsection{Real robot -- Interactive game demonstration}
\label{subsec:bubbles}
To demonstrate the operation of \textsl{HARMONIOUS} at scale and in a genuine unstructured human-robot interaction scenario, we taught the robot to play a children's board game called Bubbles\footnote{A description of the game is available at \url{https://www.piatnik.com/en/games/board-games/children-games/bubbles}. For our purposes, we simplified the game, with only six cards, three dice, only the human rolling the dice, and without grasping the cards (only pointing at the selected one). The robot recognizes the cards but is told the numbers on the dice after the roll (as these are too small to be recognized).}. Based on the numbers that appear on the dice and their colors, the players---the robot and the human in our case---compete to pick the card on the table where the size of the colored bubbles corresponds to the colors of the dice, ordered by the numbers -- see Fig.~\ref{fig:intro_photo} and video S5 in Multimedia Materials. This game was chosen because there is a shared physical space with unstructured interaction. The robot processes the visual inputs and recognizes the individual cards on the table. After the roll of the dice, the player who first places his hand over the right card wins the round. After choosing the appropriate card, the robot selects which hand to use and initiates the reach movement. At all times and in real time, visual, proximity, and tactile streams are simultaneously processed, spawning obstacles if appropriate and dynamically adding constraints to the robot controller in order to warrant the safety of the human player at all times. The game itself is rather challenging for the human player and hence in order to compete with the robot, he often cheats by pushing the robot physically away or by blocking certain volumes of the workspace by placing his hands over it. This was a deliberate choice to put \textsl{HARMONIOUS} at test.

The controller runs in dual-arm mode during the game. Depending on the position of the card in every round, the left or right arm is chosen and set as ``primary' (and hence its position reaching target is set as a constraint). Additional constraints were added to the optimization problem by adding static obstacle points (see Sec.~\ref{subsec:static_obs}) to prevent the robot from colliding with the table. We prepared a high-level Python script to play the game and used the gaze controller \cite{Roncone2016RSS} to always look at the table. The program loop starts with the processing of rolled dice (numbers sent to the program in the command line) and the current RGB-D image, followed by the detection of cards on the table, the decision of the correct card and computing its 3D position. Then one arm approaches the card position, while the other arm goes to a specific position outside the field of view of the RGB-D camera. After the reach approach (successful or not), the approaching arm returns to a specific position to clear the view for the next round. 

The robot with \textsl{HARMONIOUS} running successfully managed to run the game and cope with the unstructured interaction, dynamically processing a large number of obstacles perceived through three different sensory modalities, without endangering the human player.

\section{Conclusion, Discussion and Future work}
We have developed a real-time reactive motion control system for upper body control of a dual-arm robot with a common torso.
The proposed solution (\textsl{HARMONIOUS}) is designed to enable safe and effective interaction with humans in close proximity. 
In the absence of obstacles, we systematically tested our controller's reach ability on a grid of position and orientation targets in the workspace and compared it to \textsl{NEO} \cite{haviland_neo_2021} and \textsl{React} \cite{nguyen_merging_2018,nguyen_compact_2018}, demonstrating superior performance. We found that assigning different weights to position and orientation tasks is advantageous (unlike in \cite{haviland_neo_2021}). Preferring arm over torso movements by adjusting weights in the minimization leads to more natural-looking movements.
Uniquely to our solution, we commanded both arms and the torso of the iCub humanoid robot (in total $17$ DoFs) and demonstrated both the setup where every arm has a separate task and one, set as ``primary arm'', commands the torso joints, and a bimanual task where the hands hold an object between them. 
Kinematic singularities are handled through velocity damping and natural kinematic postures are rewarded in the optimization problem formulation.
Smooth human-like motions with a bell-shaped Cartesian velocity profile are generated using biologically inspired local trajectory sampling. Overall, several components act synergistically to produce human-like naturally looking movements. 

The solution shows the online dynamic incorporation of obstacles perceived through three different sensory modalities and their on-the-run embedding as whole-body motion constraints into a motion controller with human-like characteristics. Two of these modalities, vision and proximity,  sense at a distance, ``pre-collision'', and the other one---tactile sensors on large areas of the robot body ($2000$ individual sensors)---is providing contact or post-collision information. The performance of \textsl{HARMONIOUS} at scale was finally demonstrated in unstructured physical human-robot interaction while playing a game.

The merit of \textsl{HARMONIOUS} lies in the ability to perceive objects in the environment, relate them to the complete surface of the robot body, and, when appropriate, transform them into constraints for the motion controller. Information about obstacles from all three modalities---sometimes overlapping, at other times complementary---was processed and transformed into a unified representation dynamically generating constraints for the whole-body motion controller. 

The number or type of sensors used in this work is by no means a requirement for the described system to work. The solution will work with less dense coverage of the perirobot space, covered by only one sensor type, for example. However, dense and multimodal coverage is advantageous for the safety of the interaction. At the same time, additional sensor types could be easily incorporated. Visual obstacles and the corresponding constraints will work for any range-based sensor. Physical contact is implemented here through tactile sensors and acts as a constraint along the contact normal. If a contact force vector was available, it could be incorporated preserving the information about the direction of the constraint (as is currently done for the visual obstacles). One limitation of our current implementation is that the obstacles perceived by one sensory modality are combined into a single motion constraint per robot body part. This was an implementation decision on the obstacle processing layer. Alternatively, multiple constraints from a single modality could be preserved, transformed into constraints, and fed into the motion controller. Or, the information about obstacles from mutliple modalities could be first combined, preserving more information about the context (see contact hypotheses fusion \cite{felip2014}) and only then transformed into motion constraints.

\textsl{HARMONIOUS} is a versatile and extendable whole-body motion controller which, however, remains at the kinematic level. The problem could be approached using robot dynamics and formulated at the level of forces and torques. This may be advantageous (solving inverse dynamics rather than inverse kinematics) and ultimately more general, but at the same time it would require accurate identification of the robot inertial parameters. In general, a ``dynamics controller'' constitutes an alternative, different approach. With \textsl{HARMONIOUS} we have shown the potential and versatility of the kinematic approach, which in our view is simpler to deploy in any robot where a kinematic model and position or velocity control are available.

The problem formulation is highly modular and both the minimization criteria (e.g. motivating preferred postures) or constraints to the quadratic program (e.g. obstacles) can be easily removed or different ones added. For example, additional task constraints were added to the problem formulation for the interactive game scenario. On the other hand, our solution guarantees strict task prioritization only between constraints (e.g., obstacle avoidance) and objective function minimization (e.g., target reaching) and not between individual tasks. To enforce strict task priority, a framework of prioritized QP-based kinematic control with a null space projection can be incorporated \cite{kanoun2009}.

Our immediate future work involves adding active gaze control such that the robot can better perceive what is relevant and, at the same time, reduce the noise in the sensory streams using gaze stabilization, for example.

\bibliographystyle{IEEEtran}
\bibliography{harmonious}

\begin{IEEEbiography}[{\includegraphics[width=1in,clip,keepaspectratio]{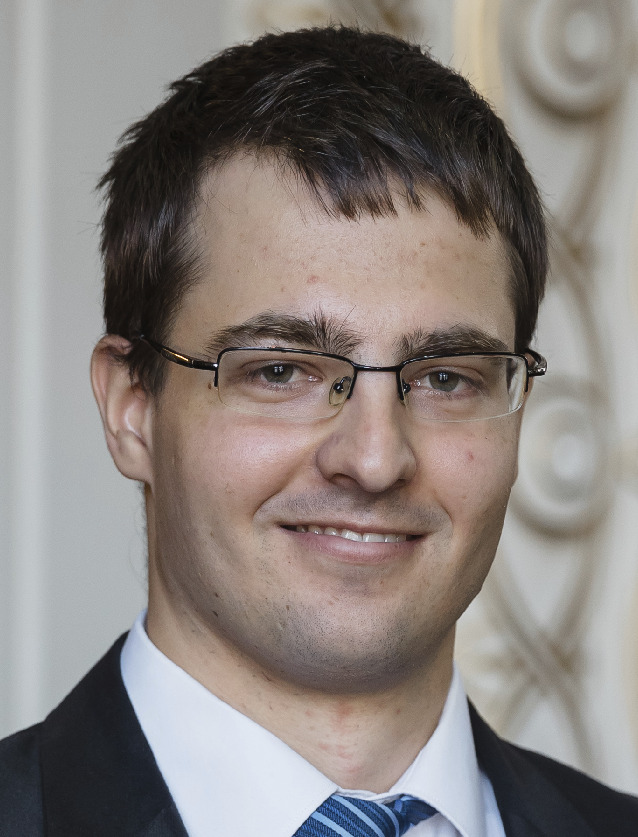}}]{Jakub Rozlivek}
is a PhD student at the Humanoid and Cognitive Robotics Laboratory (Faculty of Electrical Engineering, CTU in Prague) under Matej Hoffmann. He received his BSc. and MSc. degrees in Cybernetics and Robotics from the Faculty of Electrical Engineering, CTU in Prague in 2019 and 2022, respectively. His research interests include robot kinematic calibration, physical and social human robot interaction, robot kinematic control, and active perception.
\end{IEEEbiography}
\begin{IEEEbiography}[{\includegraphics[width=1in,clip,keepaspectratio]{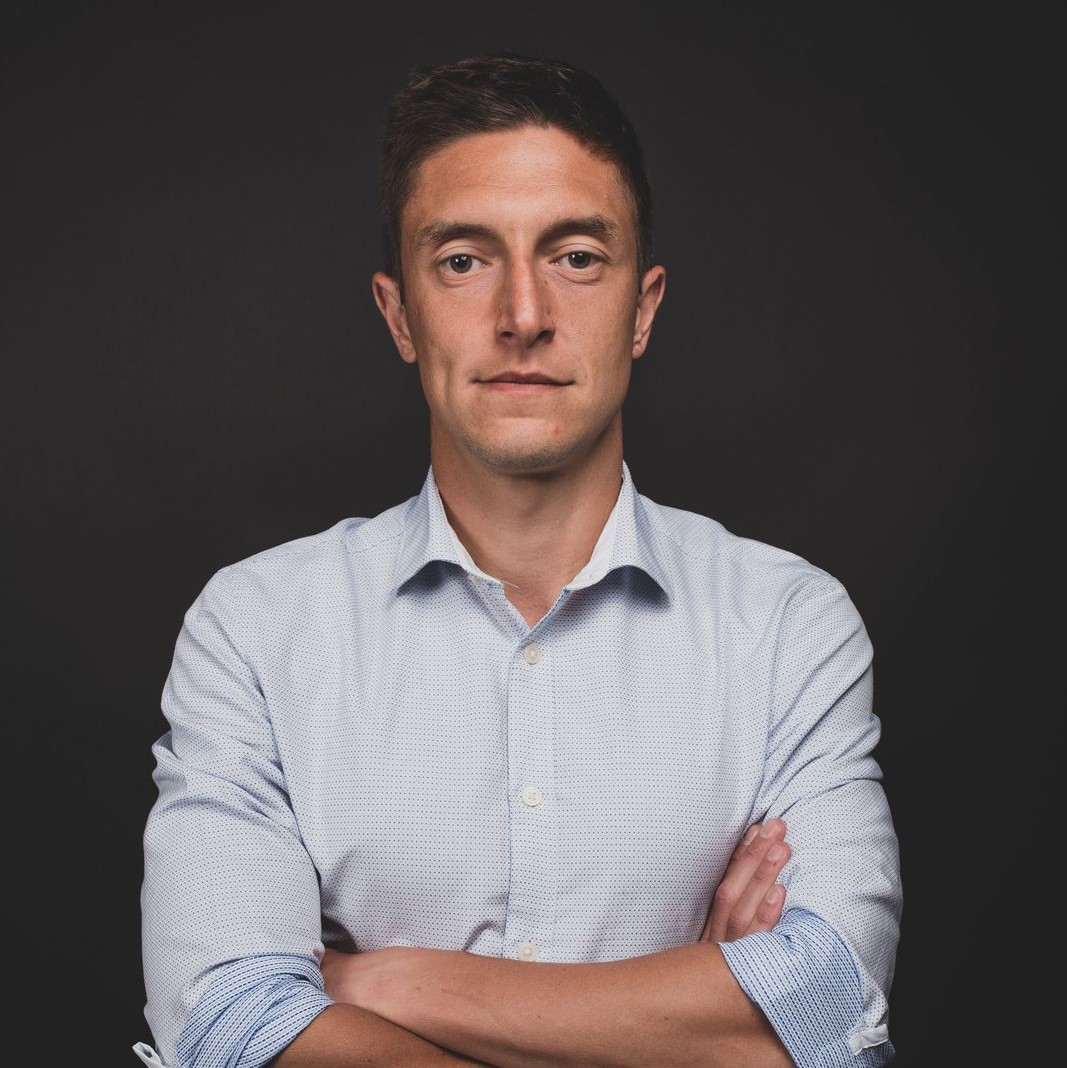}}]{Alessandro Roncone}(Senior Member, IEEE) is Assistant Professor in the Computer Science Department at University of Colorado Boulder. He received his B.Sc. \textsl{summa cum laude} in Biomedical Engineering in 2008, and his M.Sc. \textsl{summa cum laude} in NeuroEngineering in 2011 from the University of Genoa, Italy. In 2015 he completed his Ph.D. in Robotics, Cognition and Interaction Technologies from the Italian Institute of Technology [IIT], working on the iCub humanoid in the Robotics, Brain and Cognitive Sciences department and the iCub Facility. From 2015 to 2018, he was Postdoctoral Associate at the Social Robotics Lab in Yale University, performing research in Human-Robot Collaboration. He joined as faculty at CU Boulder in August 2018, where he is Director of the Human Interaction and Robotics Group (HIRO) and Associate Director of the MS/PhD Program in Robotics.
His work is at the intersection of Robotics, Artificial Intelligence, and Human-Centered Computing to design robots that help and augment human capabilities. 
\end{IEEEbiography}
\begin{IEEEbiography}[{\includegraphics[width=1in,clip,keepaspectratio]{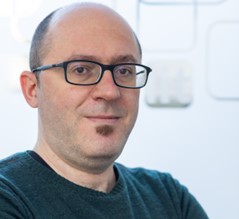}}]{Ugo Pattacini}(Member, IEEE)
is a Technologist at the Istituto Italiano di Tecnologia (IIT) of Genoa, in Italy. He graduated with honors in Electronic Engineering at the University of Pisa (2001) and he subsequently gathered 7 years of working experience in hi-tech companies in the field of Formula 1 applications (Magneti Marelli Racing Dep. and Toyota F1 Team) and aerospace (Thales Alenia Space). Ugo Pattacini received his Ph.D. in Robotics, Neurosciences, and Nanotechnologies from IIT in 2011 and he is currently involved in the development of the humanoids iCub, ergoCub, and R1, focusing on the advancement of robot motor capabilities. Ugo has also led the IIT research activities in collaboration with Fondazione Don Gnocchi ONLUS, which aims to exploit R1 as an assistive robot for rehabilitation tasks in domestic and hospital settings. Additionally, since 2023 he has been supervising the IIT team responsible for the activities of the Joint Lab with Camozzi Automation to develop smart grippers equipped with AI-enabled capabilities for autonomous and adaptive grasping. Ugo participated in the EU-funded projects RobotCub, CHRIS, EFAA, WYSIWYD, TACMAN, and ETAPAS. He is a member of IEEE, RAS, and CSS.
\end{IEEEbiography}
\begin{IEEEbiography}[{\includegraphics[width=1in,clip,keepaspectratio]{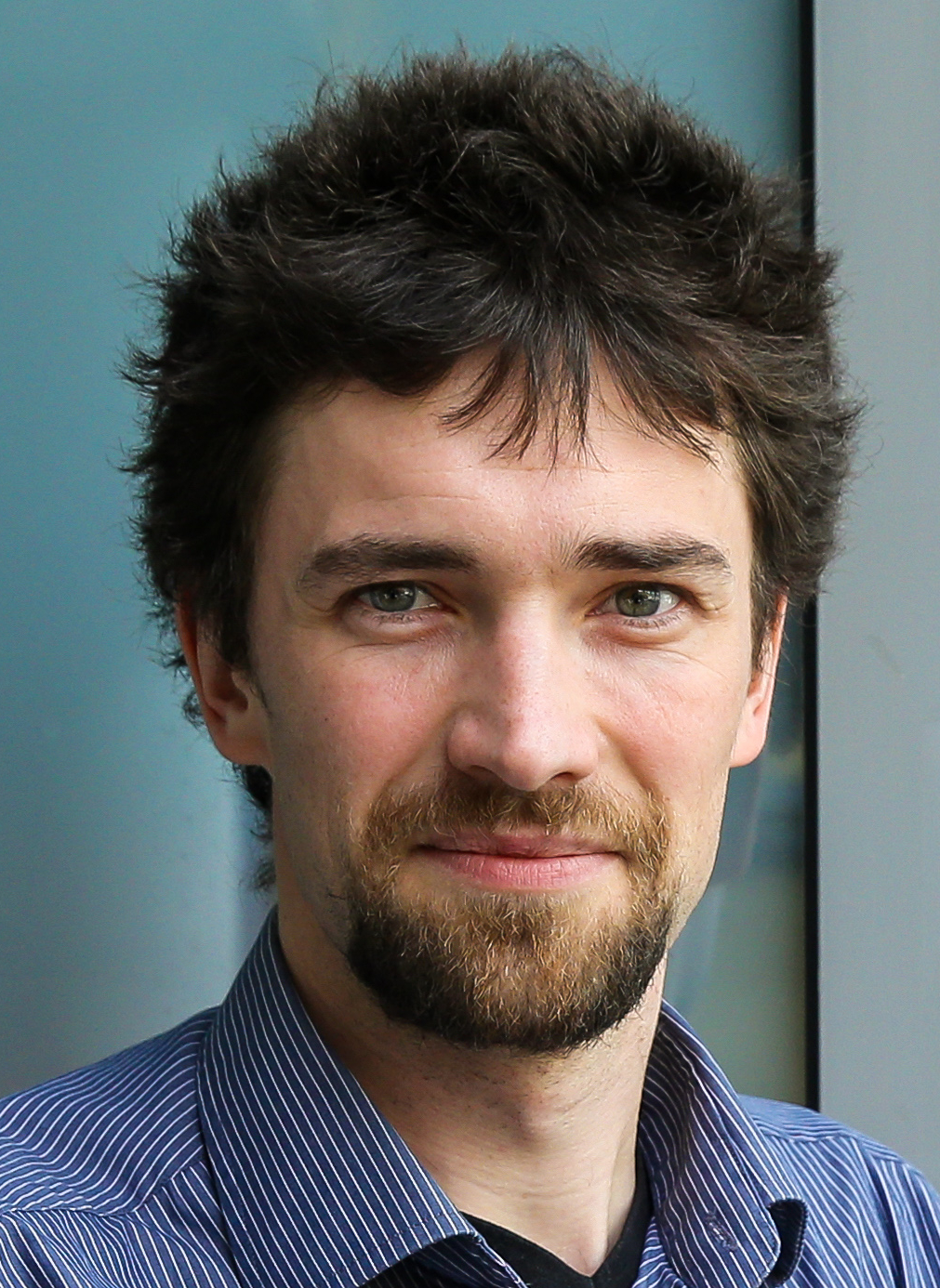}}]{Matej Hoffmann}
(Senior Member, IEEE) received the Ph.D. degree in Informatics from Artificial Intelligence Lab, University of Zurich, Zurich, Switzerland, in 2012. From 2013 to 2016, he conducted postdoctoral research with the iCub Facility of the Italian Institute of Technology, Genoa, Italy, supported by a Marie Curie Intra-European Fellowship. In 2017, he joined the Department of Cybernetics, Faculty of Electrical Engineering, Czech Technical University in Prague, where he is currently an Associate Professor and the Coordinator of the Humanoid and Cognitive Robotics Group. His research interests include humanoid, cognitive developmental, and collaborative robotics, as well as active perception for robot manipulation and grasping.
\end{IEEEbiography}

\end{document}